\documentclass[lettersize,journal]{IEEEtran}
\usepackage{amsmath,amsfonts}
\usepackage{algorithmic}
\usepackage[ruled,vlined]{algorithm2e}
\usepackage{array}
\usepackage[caption=false,font=normalsize,labelfont=sf,textfont=sf]{subfig}
\usepackage{textcomp}
\usepackage{stfloats}
\usepackage{url}
\usepackage{verbatim}
\usepackage{graphicx}
\usepackage{cite}
\usepackage{booktabs}
\usepackage{xcolor}
\usepackage{multirow} 
\usepackage{pifont}
\begin{document}

\definecolor{deepgreen}{RGB}{28,145,47}
\definecolor{darkred}{RGB}{191,15,48}
\newcommand{\cmark}{\textcolor{deepgreen}{\ding{51}}}
\newcommand{\xmark}{\textcolor{darkred}{\ding{55}}}

\title{Mobile-Aptus: Confidence-Driven Proactive and Robust \\Interaction in MLLM-based Mobile-Using Agents}

\author{Zheng Wu, Pengzhou Cheng, Zongru Wu, Yuan Guo, Tianjie Ju, Aston Zhang, \\ Gongshen Liu, Zhuosheng Zhang,~\IEEEmembership{Member,~IEEE}
		\IEEEcompsocitemizethanks{
        \IEEEcompsocthanksitem{This work is supported by Joint Funds of National Natural Science Foundation of China (U21B2020), National Natural Science Foundation of China (62406188), and Natural Science Foundation of Shanghai (24ZR1440300). Zheng Wu and Pengzhou Cheng contribute equally to this work. (Corresponding author: Zhuosheng Zhang)}
		\IEEEcompsocthanksitem{Zheng Wu, Pengzhou Cheng, Zongru Wu, Yuan Guo, Tianjie Ju, and Gongshen Liu, Zhuosheng Zhang are with the School of Computer Science, Shanghai Jiao Tong University. E-mail: (wzh815918208@sjtu.edu.cn, pengzhouchengai@gmail.com, wuzongru@sjtu.edu.cn, gy2022@sjtu.edu.cn, jometeorie@sjtu.edu.cn, lgshen@sjtu.edu.cn, zhangzs@sjtu.edu.cn).
        
        Aston Zhang is with GenAI, Meta. E-mail: (az@astonzhang.com) 
        }
        }
	}

\markboth{Journal of \LaTeX\ Class Files,~Vol.~14, No.~8, August~2021}%
{Shell \MakeLowercase{\textit{et al.}}: A Sample Article Using IEEEtran.cls for IEEE Journals}


\maketitle

\begin{abstract}
Recent advancements in multimodal large language models (MLLMs) have shown exceptional potential in enabling mobile-using agents to autonomously execute human instructions. 
However, fully automated agents often try to execute tasks even when they are unable to resolve them, leading to the problem of over-execution.
Previous studies solve it by training a interactive mobile-using agents to let agents request human interaction when agents can not complete user instructions. 
However, we find that these interactive agents tend to exhibit over-soliciting behavior, relying excessively on human intervention.
To mitigate both over-execution and over-soliciting, we propose a universal confidence integration framework that enables confidence-driven proactive and robust interaction in MLLM-based mobile-using agents.
The framework consists of two stages: interaction capability empowerment and confidence bias correction. 
In the interaction capability empowerment stage, agents learn through supervised fine-tuning to output both actions and confidence scores. 
In the confidence bias correction stage, agents learn to output more accurate confidence scores by combining semantic similarity retrieval with direct preference optimization.
Experimental results show Mobile-Aptus achieves state-of-the-art performance on the four popular mobile-using agent benchmarks: OS-Kairos, AITZ, Meta-GUI, and AndroidControl.
Mobile-Aptus consistently outperforms all baselines in offline benchmarks, with an average improvement over 17\% in task success rate.
In real-world dynamic experiments, Mobile-Aptus surpasses the baseline by 26\% in task success rate with only 0.64 intervention steps per instruction. 
Moreover, we find that the salient human interventions can be effectively replaced by a multi-agent system to avoid task interruption.
The codes are available at \url{https://github.com/Wuzheng02/Mobile-Aptus}.
\end{abstract}


\section{Introduction}
\IEEEPARstart{R}{ecently}, multimodal large language models (MLLMs) have achieved remarkable advancements in visual perception~\cite{li2023blip,wang2023image}, task planning~\cite{driess2023palm, ahn2022can}, and reasoning capabilities~\cite{liu2023visual,DBLP:journals/taslp/WangWXLF24}, laying a strong foundation for the development of mobile-using agents. 
Mobile-using agents are designed to execute human instructions within smartphones~\cite{zhang2023you,wu2024atlas,qin2025ui,ye2025mobile}.
Early mobile-using agents~\cite{steven2000jrapture,memon2003dart,memon2001hierarchical,hellmann2011rule} were primarily based on script-driven or rule-driven frameworks, which exhibited limited adaptability and flexibility. 
The latest evolution of MLLMs presents a transformative opportunity to develop more general and capable mobile-using agents based on graphical user interface (i.e., GUI agents), enabling broader applications in daily mobile automation~\cite{liu2025llm,nguyen2024gui,wang2024gui}.

\begin{figure}[t]
    \centering
    \includegraphics[width=1\linewidth]{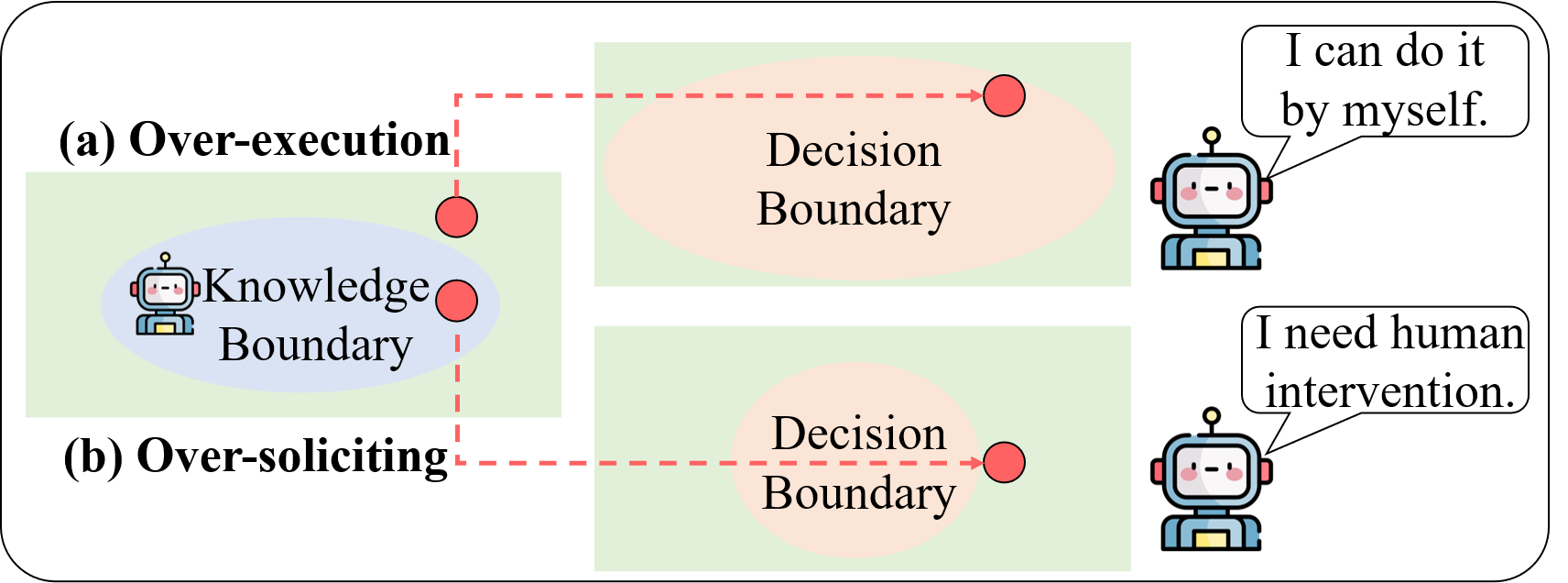}
    \caption{The decision boundary of a fully autonomous agent exceeds its actual knowledge boundary, leading to confident over-execution. In contrast, existing interactive agents have a decision boundary smaller than their actual knowledge boundary, making them prone to over-soliciting human intervention.}
    \vspace{-1em}
    \label{challenge}
\end{figure}

Recent advances have significantly improved mobile-using agents in various aspects, including grounding capability~\cite{wu2025smoothing, tang2025gui, zhou2025gui}, comprehensive perception~\cite{zhang2023you,hong2024cogagent, ma2024comprehensive}, task reasoning~\cite{zhang2024android,wang2025mobile,li2024appagent}, and self-reflection~\cite{wang2025mobileE,li2025mobileuse}. 
\begin{figure*}[t]
    \centering
    \includegraphics[width=1\linewidth]{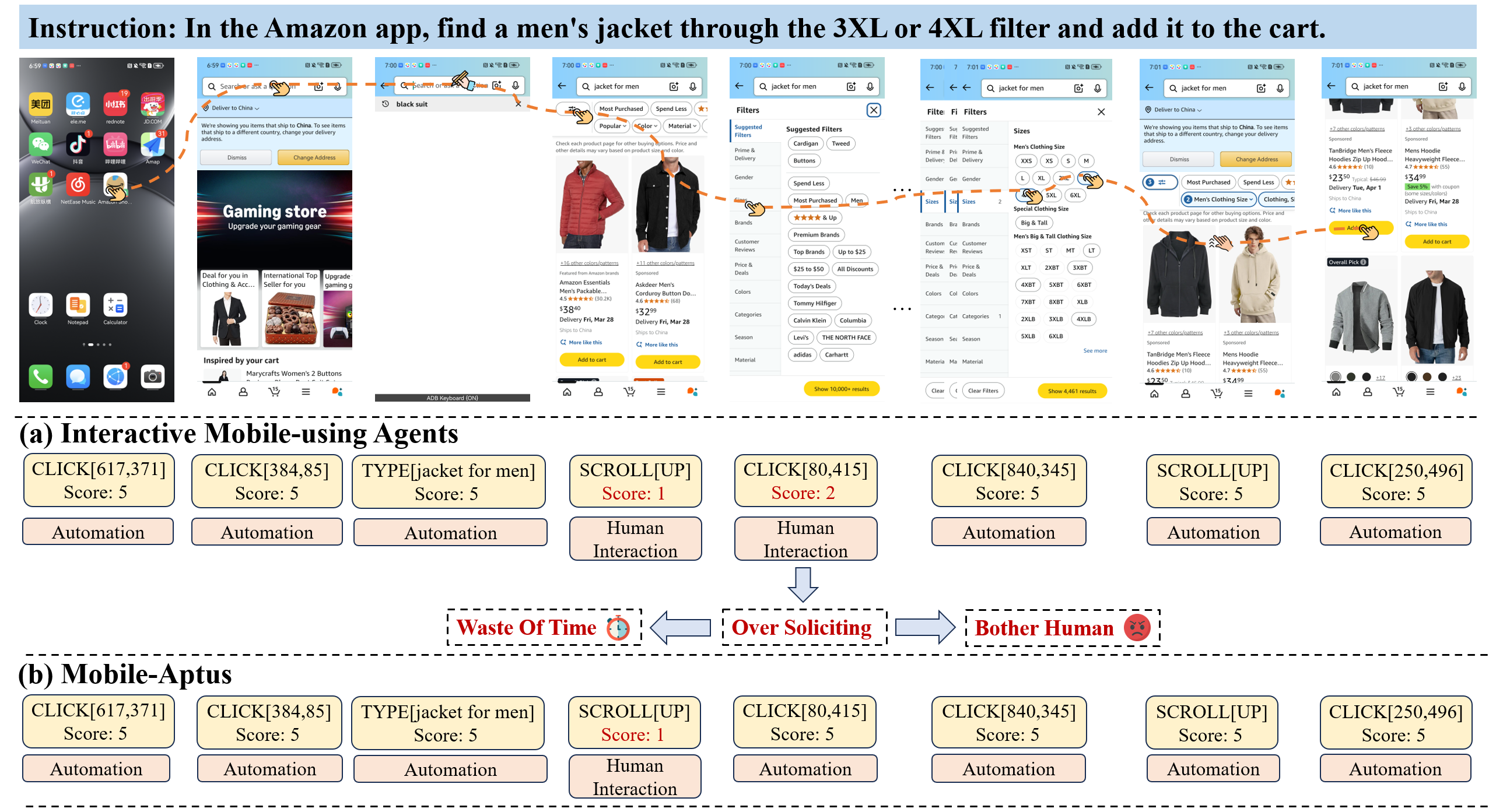}
    \vspace{-0.4cm}
    \caption{(a) Example of interactive mobile-using agents executing instructions, which resolves the over-execution issue but introduces over-soliciting. They may still request human interaction even when capable of completing tasks independently. (b) Example of Mobile-Aptus executing instructions, which addresses both over-execution and over-soliciting issues, making the timing of human intervention requests more accurate.}
    \label{intro}
    \vspace{-0.3cm}
\end{figure*}
As a result, their task success rates continue to rise. 
Despite these improvements, current research~\cite{zhang2024large,liu2025llm,shi2025towards} focuses mainly on enhancing fully autonomous execution. 
As shown in Fiugre~\ref{challenge}, recent studies~\cite{huq2025cowpilot, cheng2025kairos,wu2025verios,ou2025browseconf,subramani2025mice} observe that such agents often suffer from \textbf{over-execution} when facing unsafe scenarios~\cite{lu2025eva}, failing to assess the confidence of their actions and proceed with execution even when they should request human intervention and have explored human-agent collaboration strategies.

However, we find that these interactive agent studies may request human intervention for steps that could be automated, due to inaccurate confidence scores.
We refer to this issue as \textbf{over-soliciting}.
Over-soliciting, on one hand, leads to a waste of time and increases task execution time; on the other hand, its frequent interventions bother humans, contradicting the original purpose of mobile-using agents.

To address both over-execution and over-soliciting simultaneously, we propose a universal confidence integration framework that enables confidence-driven
proactive and robust interaction in MLLM-based mobile-using
agents. 
The framework consists of two stages: interaction capability empowerment and confidence bias correction.

In the interaction capability empowerment stage, we enable the base model to learn how to generate actions while simultaneously producing corresponding confidence scores through supervised fine-tuning (SFT) on data annotated with confidence scores (e.g., the OS-Kairos dataset~\cite{cheng2025kairos}).
                        
In the confidence bias correction stage, we design a DPO positive–negative sample pair construction algorithm using the same data in the previous stage.
This algorithm prioritizes learning from data where mobile-using agents exhibit poor confidence score predictions, ensuring a more effective adaptation to challenging cases. 
This enhancement allows the agent to maximize task completion while minimizing unnecessary human intervention.

To summarize, our contributions are primarily four-fold:

(i) We reveal that while automated mobile-using agents face challenges of over-execution, interactive mobile-using agents confront challenges of over-soliciting, and presents a systematic analysis of these two issues. 
We offer a novel perspective for advancing the community’s research on mobile-using agents.

(ii) We introduce Mobile-Aptus, a mobile-using agent capable of autonomously determining whether human intervention is necessary. 
It executes user instructions while mitigating the issue of over-execution and requests human intervention only when automation fails, minimizing unnecessary disruptions and addressing the over-soliciting issues.

(iii) We propose a universal confidence integration framework that consists interaction capability empowerment stage and confidence bias correction stage, allowing the agent to maximize task completion while minimizing unnecessary human intervention.

(iv) Experimental results on the dataset generated during the probing stage of four widely used mobile-using agent benchmarks demonstrate that Mobile-Aptus outperforms existing state-of-the-art agents. 
It achieves an average improvement of 17.47\% in task success rate. 
Moreover, when human interventions are replaced with a multi-agent system built on GPT-4o to avoid task interruption, Mobile-Aptus still yields a 32.91\% improvement in task success rate over the baseline.

\section{Related Work}

We first introduce the historical evolution of mobile-using agents.
Then we provide an analysis between the existing interactive mobile-using agents and Mobile-Aptus.

\subsection{Mobile-using Agent}

Mobile-using agents are a type of agent designed to assist humans in operating the smartphones. 
Mobile-using agents can be classified into four categories: script-based mobile-using agents, rule-based mobile-using agents, API-based mobile-using agents, and GUI-based mobile-using agents.

Early work attempted to build mobile-using agents using script-based or rule-based approaches. 
Hellman et al.~\cite{hellmann2011rule} proposed a method combining manual exploratory testing and rule-based automated testing for GUI operations. 
Steven et al.~\cite{steven2000jrapture} proposed a solution using XML-based analysis rules for operating virtual machines.

However, these work relied heavily on script-based or rule-based logic, lacking flexibility. 
As a result, some work attempted to build GUI-based mobile-using agents to enhance the flexibility of mobile-using agents. 
GUI mobile-using agents can execute user commands by performing operations such as clicking, swiping, and other actions similar to human behavior. 
Generally, GUI mobile-using agents can be divided into prompting-based GUI mobile-using agents and open-source-based GUI mobile-using agents. 
Prompting-based GUI mobile-using agents are built using closed-source large language models like GPT-4. 
For example, Mobile-Agent-V2~\cite{wang2025mobile} employed different prompts to invoke GPT-4 in various roles, constructing a multi-agent system to accomplish tasks. 
AppAgent-v2~\cite{li2024appagent} proposed a two-phase framework consisting of exploration and deployment, where the deployment phase leverages the knowledge accumulated during the exploration phase through RAG to enhance the ability of mobile-using agents to learn from historical experiences and improve instruction execution. 
However, prompt-based GUI mobile-using agents are dependent on closed-source models, which not only result in slower speeds but also incur unnecessary costs for users. 
In contrast, open-source-based GUI mobile-using agents present a more cost-effective solution, enabling GUI operation capabilities through pre-training, supervised fine-tuning (SFT) or RL methods applied to open-source models. 
Some work~\cite{zhang2023you,ma2024comprehensive,hong2024cogagent,xu2024aguvis,wu2024atlas,ye2025mobile,qin2025ui} provides a domain model for GUI mobile-using agents.
Some works have also explored RL for mobile-using agents. 
For example, DigiRL~\cite{zhoudigirl}, DistRL~\cite{wang2024distrl}, UI-R1~\cite{lu2025ui}, GUI-R1~\cite{luo2025gui}, InfiGUI-R1~\cite{liu2025infigui} and UI-S1~\cite{lu2025uis1} explored RL to improve agents' execution of instructions in real-world environments.

Despite the rapid advancement of mobile-using agents, they still mainly run in a fully automated manner and continue to struggle with complex tasks and unexpected environmental variations, leading to the issue of over-execution. 
If mobile-using agents could request human intervention at some moments when they require assistance, it would significantly enhance their practical usability.

\subsection{Interactive Mobile-using Agents}

Extensive research on mobile-using agents emphasizes their capacity for autonomous task execution. 
However, since these agents cannot independently handle all possible human commands, facilitating human-agent collaboration can further enhance their effectiveness. 
CowPilot~\cite{huq2025cowpilot} identifies this issue but relies on manually determining intervention timing. 

Some work attempts to address the issue of over-execution by building interactive mobile-using agents.
For example, OS-Kairos~\cite{cheng2025kairos} employs a confidence-driven approach, enabling agents to output a confidence score and request human intervention when the score falls below a threshold. 
VeriOS~\cite{wu2025verios} and InquireMobile~\cite{ai2025inquiremobile} utilize a query-driven method to seek human assistance when intervention is required.

However, these work may request human intervention for steps that could be automated, due to inaccurate confidence judgement, leading to the issue of over-soliciting.
Mobile-Aptus is an interactive mobile-using agent that has more accurate confidence judgement, addressing both over-execution and over-soliciting.


\section{Motivation}

We first introduce the working paradigms of fully automated mobile-using agents and interactive mobile-using agents. 
Then, we describe the over-soliciting problem encountered when incorporating human-agent interaction and discuss its potential impacts.

\subsection{Working Paradigms}
In the following, we formally introduce the working paradigms of both fully automated mobile-using agents and adaptive interaction mobile-using agents, respectively.

\textbf{Fully Automated mobile-using agents.} Given a fully automated mobile-using agent $\mathcal{F}$ and a user instruction $\tau$, the agent starts from the initial screenshot $s_0$ and infers the first action $a_0$ required to accomplish $\tau$. 
It then executes $a_0$ in the environment and appends it to the action history $h_0$.
This process repeats iteratively until $\tau$ is completed. 

Formally, the working paradigm of a fully automated mobile-using agent $\mathcal{F}$ can be defined as follows: at step $t$ of execution, $\mathcal{F}$ predicts the action $a_t$ as
\begin{equation}
a_t = \mathcal{F}( s_t, h_{t-1}, \tau),
\end{equation}
where $a_t$ will be executed in the environment, resulting in the next screenshot $s_{t+1}$. This continues until the screenshot $s_{t+1}$ satisfies the completion criteria for $\tau$.

\textbf{Adaptive Interaction mobile-using agents.}  
Given an adaptive interaction mobile-using agent $\mathcal{F}$, a user instruction $\tau$, and a user-defined intervention threshold $\gamma$, the agent starts from the initial screenshot $s_0$ and infers both the first action $a_0$ and its confidence score $c_0$. If $c_0 > \gamma$, the agent executes $a_0$, interacts with the environment to obtain the new state $s_1$, and appends $a_0$ to the action history $h_0$. Otherwise, if $c_0 \leq \gamma$, $\mathcal{F}$ requests human intervention. The human provides a new action $a'_0$, which is then executed, leading to the new state $s_1$, with $a'_0$ appended to the action history.  This process repeats iteratively until $\tau$ is completed.

Formally, the working paradigm of an adaptive interaction agent $\mathcal{F}$ can be defined as follows: at each step $t$, the agent infers an action $a_t$ with a confidence score $c_t$ as
\begin{equation}  
    (a_t, c_t) = \mathcal{F}(s_t, h_{t-1}, \tau), 
\end{equation} 

If the confidence score surpasses the threshold $\gamma$, the action is executed autonomously; otherwise, human assistance is requested. Thus, we have
\begin{equation}  
    a_t^* =  
    \begin{cases}  
        a_t, & \text{if } c_t > \gamma, \\  
        a'_t, & \text{if } c_t \leq \gamma  
    \end{cases}  
\end{equation}  
where $a_t^*$ will be executed in the environment, resulting in the next screenshot $s_{t+1}$. This continues until the screenshot $s_{t+1}$ satisfies the completion criteria for $\tau$.

\subsection{Challenge of Over-soliciting}

When agents are capable of assessing confidence in their actions and requesting human intervention when necessary, it is equally important to ensure they do not do so excessively. 
Ideally, an intelligent agent should strike a balance between self-sufficiency and appropriate reliance on human input. However, as illustrated in Figure \ref{intro}, interactive mobile-using agents may sometimes assign a low confidence score to actions that are actually correct, resulting in unnecessary human intervention requests. 
This miscalibration in confidence estimation leads to a phenomenon we define as over-soliciting.

Over-soliciting poses several challenges. 
Frequent and unnecessary human interventions not only waste users’ time but also lead to cognitive overload and frustration, ultimately diminishing the overall efficiency and user experience. 
Moreover, excessive reliance on human input undermines the core goal of mobile-use agents: to automate tasks rather than add user burden.
If left unaddressed, this issue can erode trust in the system, as users may perceive the agent as unreliable or overly dependent on their guidance.

To mitigate the problem of over-soliciting, our work introduces an improved framework, Mobile-Aptus. 
This enhanced model is designed to refine confidence estimation, reducing unnecessary intervention requests while preserving the ability to seek human input when truly needed. 
By balancing autonomy with selective human collaboration, Mobile-Aptus enables a proactive and robust interaction paradigm. 

\section{Methodology}

\begin{figure*}[!htbp]
    \centering
    \includegraphics[width=1\linewidth]{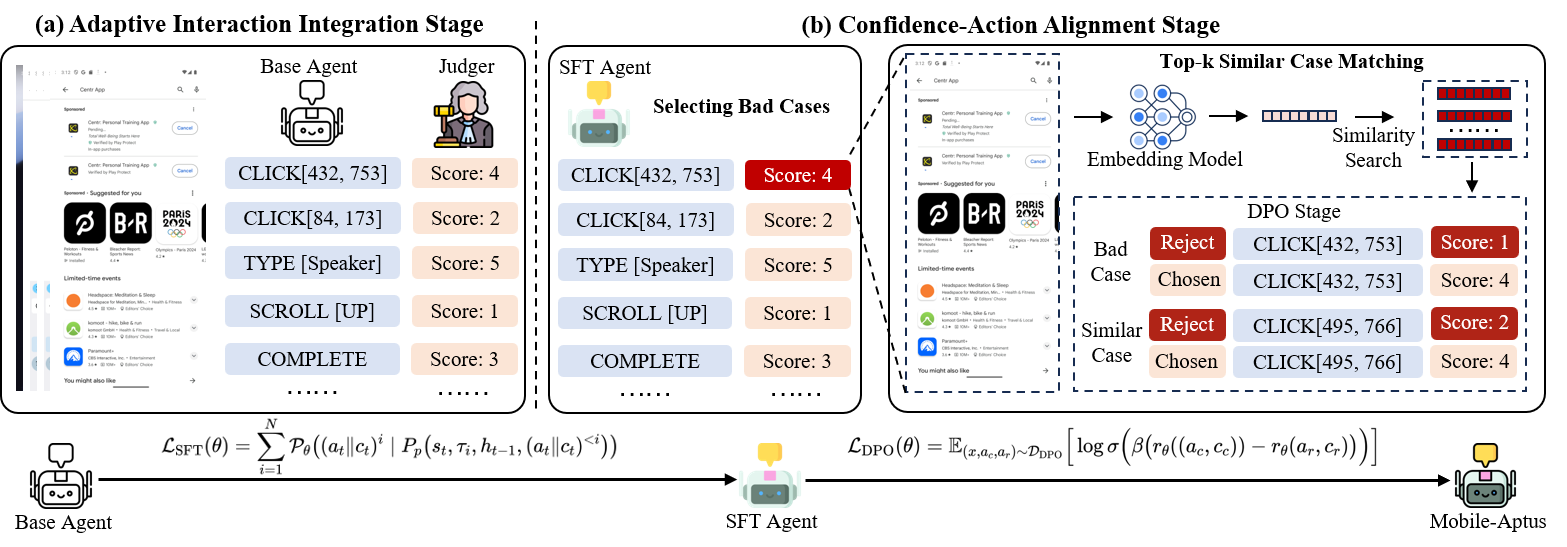}
    \vspace{-0.4cm}
    \caption{Overview of the universal confidence integration framework. First, leverage the data that has ground-truth action and confidence score, we apply SFT to enable the probed mobile-using agent to generate confidence scores alongside action predictions without altering its ability to generate actions. 
    Then, we employ a confidence bias correction strategy based on semantic similarity retrieval and DPO to construct a DPO dataset, further improving the accuracy of the agent’s confidence score generation. }
    \label{pipeline}
    \vspace{-0.3cm}
\end{figure*}

This section provides an overview of our universal confidence integration framework. 
Our goal is to design a framework that enables a mobile-using agent to autonomously determine whether human intervention is necessary. 
Figure ~\ref{pipeline} overviews the pipeline of our methodology.

This pipeline consists of two stages: the adaptive interaction integration stage and the confidence bias correction stage.
In the adaptive interaction integration stage, we applied SFT to equip the agent with adaptive interaction integration. 
In the confidence bias correction stage, we introduced a confidence bias correction strategy based on semantic similarity retrieval and DPO, which effectively reduces unnecessary intervention requests while maintaining robust interaction performance.


\subsection{Adaptive Interaction Integration Stage}
At this stage, we integrate confidence scoring through SFT to produce a mobile-using agent capable of interaction.

We have obtained the confidence score \(\text{score}_t\) for the action \( a_t \) generated by the mobile-using agent \( \mathcal{F}\) using the OS-Kairos~\cite{cheng2025kairos} method to give \( a_t \) a confidence score.
In particular, the confidence score is inherently agent-dependent; given the same instruction and screenshot, different agents may assign different confidence scores. Therefore, this dataset annotated with confidence scores can, to some extent, reflect the capability boundaries of different agents.
We refer to this set of data as \(\mathcal{D}_{\text{SFT}}\).

In SFT stage, we enable \( \mathcal{F}\) to autonomously generate \(\text{score}_t\) along with \( a_t\). 
By comparing \(c_t\) with the user-set threshold \( \gamma \), the agent can autonomously determine whether human intervention is required.

Since there is a one-to-one correspondence between \(c_t\) and \( a_t \) in \(\mathcal{D}_{\text{SFT}}\), to ensure the validity of \(c_t\), our objective is to ensure that the action output by \(\mathcal{F}\) aligns with \(a_t\) in \(\mathcal{D}_{\text{SFT}}\).
This design ensures that \( \mathcal{F} \) learns to generate \(c_t\) without altering its ability to produce \( a_t\).

Formally, the joint probability of predicting the action \(a_t\) and confidence score \(c_t\) conditioned on the state \(s_t\), task description \(\tau_t\), and history \(h_{t-1}\) can be factorized as:
\begin{equation}
    \begin{split}
    \mathcal{P}_\theta((a_t||c_t) \mid s_t, \tau_t, h_{t-1})
    = \prod_{i=1}^{N_a} \mathcal{P}_\theta(a_t^i \mid a_t^{<i}, s_t, \tau_t,\\ h_{t-1})\times \prod_{j=1}^{N_c} \mathcal{P}_\theta(c_t^j \mid c_t^{<j}, a_t, s_t, \tau_t, h_{t-1}),
    \end{split}
\end{equation}
where \(N_a\) and \(N_c\) represent the token lengths of action \(a_t\) and confidence score \(c_t\), respectively.

Accordingly, the training objective \(\mathcal{L}_{\text{SFT}}\) for next-word prediction can be expressed as:
\begin{equation}
    \begin{aligned}
    \mathcal{L}_{\text{\textit{SFT}}}(\theta) & =\sum_{i=1}^N \mathcal{P}_\theta((a_t||c_t)^i \mid  \\ &P_p(s_t, \tau_i, h_{t-1}, (a_t||c_t)^{<i})),
    \end{aligned}
\end{equation}
where \( N \) represents the number of tokens in \( a_t \) and \( c_t \), and \( || \) denotes the concatenation operator for action prediction and score.
The trainable parameters in \( \mathcal{F}_p \)  are represented by \( \theta \). 
This optimization approach offers greater stability compared to multi-task learning, as it not only maintains \( \mathcal{F}\) ’s capability to predict actions but also generates confidence scores for the predicted actions. 
The model obtained from this stage is denoted as \( \mathcal{F}_{SFT}\) .

\subsection{Confidence Bias Correction Stage}
\begin{algorithm}[t]
\caption{Confidence Bias Correction Stage}
\label{alg:dpo-stage}
\SetAlgoLined
\DontPrintSemicolon
\SetKwInOut{Require}{Require} 
\SetKwInOut{Ensure}{Ensure}   

\Require{
    \( \mathcal{F}_{SFT}\), text encoder $E_t$, vision encoder $E_v$, SFT dataset $\mathcal{D}_{\text{SFT}}$, intervention threshold $\gamma$, similarity threshold $\lambda$, top-$k$ value
}
\Ensure{
    DPO training dataset $\mathcal{D}_{\text{DPO}}$, optimized model Mobile-Aptus
}

$\mathcal{D}_{\text{DPO}} \gets \emptyset$\;

\For{each $(x, a_{\text{SFT}}, c_{\text{SFT}}) \in \mathcal{D}_{\text{SFT}}$}{
    Encode $x$ using $E_t$ and $E_v$ to get feature vector $v_x$\;
}

\For{each $(x, a_{\text{SFT}}, c_{\text{SFT}}) \in \mathcal{D}_{\text{SFT}}$}{
    Obtain $(a, c) = \mathcal{F}_{SFT}(x)$\;
    \If{$(c_{\text{SFT}} \geq \gamma \land c \geq \gamma) \lor (c_{\text{SFT}} < \gamma \land c < \gamma)$}{
        \textbf{continue}\;
    }
    \For{each $(x', a'_{\text{SFT}}, c'_{\text{SFT}}) \in \mathcal{D}_{\text{SFT}}$}{
        Encode $x'$ to get feature vector $v_{x'}$\;
        Compute cosine similarity between $x$ and $x'$: \\
        \quad $\text{sim}(x, x') = \frac{\langle v_x, v_{x'} \rangle}{\|v_x\| \cdot \|v_{x'}\|}$\;
    }
    Select top-$k$ samples $x'$ with highest $\text{sim}(x, x')$\;
    \For{each selected $(x', a'_{\text{SFT}}, c'_{\text{SFT}})$}{
        Obtain $(a', c') = \mathcal{F}_{SFT}(x)$\;
        \If{$\text{sim}(x, x') > \lambda$}{
            \If{$(c'_{\text{SFT}} \geq \gamma \land c' \geq \gamma) \lor (c'_{\text{SFT}} < \gamma \land c' < \gamma)$}{
                $c'\gets \arg\max_{s \in \{1,2,3,4,5\}} |s - c'_{\text{SFT}}|$\;
            }
            Construct sample pair: \\
            \quad $\text{chosen} = (a'_{\text{SFT}}, c'_{\text{SFT}})$,
            $\text{rejected} = (a', c')$\;
            $\mathcal{D}_{\text{DPO}} \gets \mathcal{D}_{\text{DPO}} \cup \{(x', \text{chosen}, \text{rejected})\}$\;
        }
    }
}

\For{each training step on $\mathcal{D}_{\text{DPO}}$}{
    Sample $(x, (a_c, c_c), (a_r, c_r))$\;
    Compute rewards: \\
    \quad $r_\theta((a, c)) = \log \pi_\theta((a, c) \mid x) - \log \pi_0((a, c) \mid x)$\;
    Compute DPO loss: \\
    \quad $\mathcal{L}_{\text{DPO}} = -\log \sigma \left( \beta \left( r_\theta((a_c, c_c)) - r_\theta((a_r, c_r)) \right) \right)$\;
    Update $\theta$ via gradient descent on $\mathcal{L}_{\text{DPO}}$\;
}

\Return{$\mathcal{D}_{\text{DPO}}$, Mobile-Aptus}
\end{algorithm}

At this stage, we propose
a confidence bias correction strategy that combines semantic similarity retrieval with DPO~\cite{rafailov2023direct} to mitigate the challenge of over-soliciting. 
The entire stage is described in Algorithm~\ref{alg:dpo-stage}.

\subsubsection{\textbf{Semantic Similarity Retrieval}}
Given the SFT dataset $\mathcal{D}_{\text{SFT}}$, we first encode each input $x$ using the text encoder $E_t$ and vision encoder $E_v$, obtaining its feature representation $\mathbf{v}_x$. 
For each sample $(x, a_{\text{SFT}}, c_{\text{SFT}})$ in $\mathcal{D}_{\text{SFT}}$, we use \( \mathcal{F}_{SFT}\) to generate its corresponding action and confidence score $(a, c)$.

If both $c$ and $c_{\text{SFT}}$ are either above or below the intervention threshold $\gamma$, the sample is skipped. 
Otherwise, we compute the cosine similarity between the input $x$ and each other sample $x' \in \mathcal{D}_{\text{SFT}}$ based on their feature representations. These representations, denoted as $\mathbf{v}_x$ and $\mathbf{v}_{x'}$, are obtained by encoding the input through both the text encoder $E_t$ and the vision encoder $E_v$. The cosine similarity is defined as:  
\begin{equation} 
\text{sim}(x, x') = \frac{\langle \mathbf{v}_x, \mathbf{v}_{x'} \rangle}{\|\mathbf{v}_x\| \cdot \|\mathbf{v}_{x'}\|}. 
\end{equation} 

After computing the cosine similarity between the input sample $x$ and all other samples $x' \in \mathcal{D}_{\text{SFT}}$, in order to provide \( \mathcal{F}_{SFT}\) with more information to learn from, we proceed to identify the most relevant examples by performing a top‑$k$ selection. 

Then, among these \( k \) samples, if the similarity is greater than the similarity threshold \( \lambda \) and \( c'\) is incorrect, we construct \( (a'_{\text{SFT}}, c'_{\text{SFT}}) \) as the chosen sample for DPO and \( a', c'\) as the rejected sample. 
However, if \( c'\) is already a correct confidence score, to satisfy DPO's requirement for constructing negative samples, we generate an incorrect confidence score \( c' \) using
\begin{equation}
c' = \arg\max_{c \in \{1,2,3,4,5\}} |c - c'_{\text{SFT}}|
, 
\end{equation}
and then construct \( (a', c') \) as the rejected sample.
Then, we use the collected chosen and rejected sample pairs to construct triplets \((x', \text{chosen}, \text{rejected})\) and store them as \(\mathcal{D}_{\text{DPO}}\).

\subsubsection{\textbf{DPO stage}}
After constructing $\mathcal{D}_{\text{DPO}}$, we optimize the \( \mathcal{F}_{SFT}\) model based on $\mathcal{D}_{\text{DPO}}$. 
The core idea is to directly fine-tune \( \mathcal{F}\) so that it learns to favor responses requesting human intervention at appropriate moments, while discouraging those where the intervention occurs at inappropriate times.

The reward function is defined as follows:
\begin{equation}
r_\theta((a, c))= \log \pi_\theta((a, c) \mid x) - \log \pi_0((a, c) \mid x),
\end{equation}
where $\pi_\theta$ represents the current model’s policy, and $\pi_0$ is the initial policy. The DPO loss function is then optimized:
\begin{equation}
\begin{split}
\mathcal{L}_{\text{DPO}}(\theta) = 
\mathbb{E}_{(x, a_c, a_r) \sim \mathcal{D}_{\text{DPO}}} \Big[
\log \sigma \Big( \beta \big( r_\theta((a_c, c_c)) \\
- r_\theta(a_r, c_r) \big) \Big) \Big],
\end{split}
\end{equation}
where $\beta$ is a hyperparameter that scales the reward values. 
By minimizing this loss, \( \mathcal{F}_{SFT}\)  will make more accurate judgments on the confidence score, reducing the frequency of requesting intervention, thereby addressing the over-soliciting issues. 
Finally, we obtain the optimized model Mobile-Aptus.

\begin{table}[ht]
    \centering
    \small
    \caption{Dataset statistics of the evaluated benchmarks.}
    \label{dataset_statistics}
    \begin{tabular}{lccc}
     \toprule
     \textbf{OS-Kairos} & \textbf{Trajectory} & \textbf{Screen} & \textbf{Goal} \\ \midrule
      Train & 800 & 4078 & 759  \\ 
      Test & 200 & 1054 & 198 \\ \midrule
      
      \textbf{AITZ} & \textbf{Trajectory} & \textbf{Screen} & \textbf{Goal} \\ \midrule
      General & 479 & 3607 & 479 \\
      Install & 420 & 3627 & 420 \\ 
      Google Apps & 242 & 1889 & 242 \\
      Single & 844 & 2594 & 844 \\
      Web Shopping & 519 & 6926 & 519 \\ \midrule
      
      \textbf{Meta-GUI} & \textbf{Trajectory} & \textbf{Screen} & \textbf{Goal} \\ \midrule
      Train & 897 & 14539 & 2286 \\
      Test  & 116 & 1923 & 336 \\ \midrule
      
      \textbf{AndroidControl} & \textbf{Trajectory} & \textbf{Screen} & \textbf{Goal} \\ \midrule
      Train &13593  &74714  &12950  \\
      Test  &1543  &8444  &1524  \\ \bottomrule
    \end{tabular}
\end{table}

 \begin{table*}[ht]
    \centering
    \small
    \caption{Action Space In Our Experiment.}
    \label{action_space}
    \begin{tabular}{ccc}
     \toprule
    \textbf{Action Type} & \textbf{Action Description} & \textbf{Action Format} \\ 
    \midrule 
    
    CLICK & Click at the specified position. & CLICK \textless point\textgreater[[x-axis, y-axis]]\textless/point\textgreater \\
    TYPE & Enter specified text at the designated location. & TYPE [input text] \\
    SCROLL & Scroll in the specified direction. & SCROLL [UP/DOWN/LEFT/RIGHT] \\
    PRESS\_BACK & Press a back button to navigate to the previous screen. & PRESS\_BACK \\
    PRESS\_HOME & Press a home button to navigate to the home page. & PRESS\_HOME \\
    ENTER & Press the enter button. & ENTER \\
    OPEN\_APP & Open the specified application. & OPEN\_APP [app\_name] \\
    WAIT & Wait for the screen to load. & WAIT \\
    LONG\_PRESS & Long press at the specified position. & LONG\_PRESS \textless point\textgreater[[x-axis, y-axis]]\textless/point\textgreater \\
    COMPLETE & Indicate the task is finished. & COMPLETE \\
    IMPOSSIBLE & Indicate the task is impossible. & IMPOSSIBLE \\
    \bottomrule
    \end{tabular}
\end{table*}
\section{EXPERIMENTS}
In this section, we first present the experimental setup, followed by the main results and discussion. We then provide additional experiments and analyses to further validate the effectiveness of our method. Finally, we present and analyze the results of the ablation study.
The base mobile-using agent $\mathcal{F}$ is OS-Atlas-pro-7B in our experiments.

\subsection{Experiment Setup}

\noindent\textbf{Datasets.}
We conducted experiments on multiple benchmarks, including OS-Kairos~\cite{cheng2025kairos}, AITZ~\cite{zhang2024android}, Meta-GUI~\cite{sun2022meta}, and the AndroidControl~\cite{li2024effects} benchmarks.
The introduction of the three benchmarks is as follows:
\begin{itemize}
    \item \textbf{OS-Kairos}:OS-Kairos is a dataset spanning over ten task scenarios and more than ten applications. Its distinctive feature is the annotation of confidence scores for the actions of OS-Atlas-Pro-7B. For each frame of images, this not only provides the ground truth action but also includes knowledge about the capability boundaries of OS-Atlas-Pro-7B, offering valuable data for subsequent research.
    \item \textbf{AITZ}: AITZ is a benchmark constructed based on the AITW~\cite{rawles2024androidinthewild} benchmark after data cleaning, deduplication, and filtering, enhanced with the chain-of-action thought (CoAT) technique. It contains 18,643 screen-action pairs along with CoAT annotations and is divided into five subsets: General, Install, GoogleApps, Single, and WebShopping.
    \item \textbf{Meta-GUI}: Meta-GUI is a mobile-using agents benchmark that includes 11 applications across six topics: weather, calendar, search, text, restaurant, and hotel. It covers seven common GUI operations and over a thousand task execution episodes.
    \item \textbf{AndroidControl}: AndroidControl is a large-scale dataset containing approximately 15,000 human demonstrations of task execution in Android applications. For each task, it provides both high-level and low-level human-generated instructions. It is the most diverse UI control dataset to date, encompassing unique tasks across 833 different Android applications.

\end{itemize}

Table~\ref{dataset_statistics} presents the statistics of our experiment's dataset. 
The OS-Kairos benchmark includes the confidence score annotated for OS-Atlas-Pro-7B. 
For the Meta-GUI, AITZ, and AndroidControl benchmarks, we also obtained confidence scores using the OS-Kairos method.

\noindent\textbf{Baselines.}
We have compared a wide range of baselines, including prompting-based mobile-using agents and open-source-based mobile-using agents. 
For prompting-based mobile-using agents, we conducted experiments using GPT-4o~\cite{hurst2024gpt}, GLM-4V-Plus~\cite{glm2024chatglm}, and Qwen-VL-MAX~\cite{bai2023qwen} as baselines through API calls. 
To assist these agents in recognizing coordinates on screenshots, we employed ResNet18~\cite{he2016deep} and ConvNeXt-Tiny~\cite{liu2022convnet} as OCR tools. 
For open-source-based mobile-using agents, we conducted experiments using CogAgent~\cite{hong2024cogagent}, Auto-UI~\cite{zhang2023you}, Qwen2-VL-7B~\cite{bai2023qwen}, OS-Atlas-Pro-7B~\cite{wu2024atlas}, Qwen3-VL-8B~\cite{Qwen3-VL}, UI-TARS-1.5-7B~\cite{qin2025ui} and GUI-owl-7B~\cite{ye2025mobile} as baselines.

\noindent\textbf{Metrics.}
In the experiments section, to evaluate task performance, we report the following metrics: step-wise success rate (SR) for each action type and overall, with finer-grained action type accuracy (Type) for CLICK and TYPE actions. Additionally, we report the task success rate (TSR). 
The action space presented in Table \ref{action_space} is the union of action spaces across all datasets and benchmarks, comprising 11 distinct actions that simulate possible human interventions within the operating system. 
To save table space, we represent the actions PRESS\_BACK, PRESS\_HOME, and ENTER as PRESS, and combine COMPLETE and IMPOSSIBLE as STOP. 
Additionally, the OPEN\_APP action is only present in the AndroidControl benchmark; the other three benchmarks do not include this action in their action 
We also define four statistical metrics with the user-set intervention threshold $\gamma$:
\begin{itemize}
    \item \textbf{True positive (TP)}: Both the model’s confidence score and the actual requirement are exceed $\gamma$, meaning no intervention is needed, and none is performed.
    \item \textbf{False positive (FP)}: The model’s confidence score exceeds $\gamma$, but the actual requirement does not, causing the agent to perform a missed intervention.
    \item \textbf{True negative (TN)}: Both the model’s confidence score and the actual requirement below $\gamma$, correctly leading to an intervention.
    \item \textbf{False negative (FN)}: The model’s confidence score is below $\gamma$, while the actual requirement exceeds $\gamma$, resulting in an unnecessary intervention.
\end{itemize}

To assess adaptive interaction, we define three key metrics: human intervention success rate (HSR), intervention Precision (IP), and average intervention frequency (AIF). HSR measures the accuracy of the mobile-using agent's judgment on the need for human intervention across the dataset. It is calculated as the ratio of correct predictions (both true positives and true negatives) to the total number of instances:
   \begin{equation}
       \text{HSR} = \frac{TP + TN}{TP + TN + FP + FN}.
   \end{equation}

IP quantifies the proportion of cases where human intervention is genuinely required among all instances where the model predicts the need for intervention. It is defined as:
   \begin{equation}
       \text{IP} = \frac{TN}{TN + FN}.
   \end{equation}
   
AIF is defined as the average number of requests for human intervention made by the model in an episode. In dynamic real-world evaluation, following the evaluation standards established in prior work~\cite{li2024appagent,wang2024mobile}, we additionally define the relative efficiency (RE) of the mobile-using agents compared to the steps taken by humans.

\begin{table*}[!htbp]
\centering
\renewcommand{\arraystretch}{1.2}
\setlength{\tabcolsep}{3pt}
\caption{Model Performance Comparison On OS-Kairos dataset}
\label{model_performance_on_kairos_dataset}

\begin{tabular}{lccccccccccccc}
\toprule
\multirow{2}{*}{\textbf{Models}} &
\multirow{2}{*}{\textbf{Source}} &
\multirow{2}{*}{\textbf{SCROLL}} &
\multirow{2}{*}{\textbf{PRESS}} &
\multirow{2}{*}{\textbf{STOP}} &
\multicolumn{2}{c}{\textbf{CLICK}} &
\multicolumn{2}{c}{\textbf{TYPE}} &
\multicolumn{2}{c}{\textbf{TOTAL}} &
\multirow{2}{*}{\textbf{TSR}} \\
\cmidrule(lr){6-7} \cmidrule(lr){8-9} \cmidrule(lr){10-11}
& & & & &
\textbf{Type (\%)} $\uparrow$ &
\textbf{SR (\%)} $\uparrow$ &
\textbf{Type (\%)} $\uparrow$ &
\textbf{SR (\%)} $\uparrow$ &
\textbf{Type (\%)} $\uparrow$ &
\textbf{SR (\%)} $\uparrow$ & \\
\midrule

GPT-4o          & Closed & 22.22 & \textbf{100.0} & 46.67 & 86.95 & 74.63 & 93.62 & 90.07 & 87.59 & 76.35 & 39.13 \\
GLM-4V-Plus     & Closed & 0.00  & 0.00           & 20.00 & 95.88 & 37.65 & 21.99 & 20.57 & 81.57 & 33.80 & 4.35  \\
Qwen-VL-MAX     & Closed & 0.00  & \textbf{100.0} & \textbf{92.21} & 51.25 & 38.33 & 96.45 & \textbf{92.21} & 58.73 & 46.89 & 29.81 \\

\midrule

Auto-UI         & Open   & 44.44 & 0.00 & 0.00 & 2.93 & 0.15 & 0.00 & 0.00 & 2.81 & 0.59 & 0.00 \\
Qwen2-VL-7B     & Open   & 22.22 & 85.71 & 0.00 & 37.98 & 15.69 & 55.32 & 42.55 & 40.75 & 20.49 & 0.00 \\
OS-Atlas-Pro-7B & Open   & \textbf{66.67} & 0.00 & 20.00 & 97.80 & 62.46 & 99.29 & 63.12 & 95.90 & 61.36 & 14.29 \\
Qwen3-VL-8B     & Open   & 0.00 & 0.00 & 26.67 & 95.88 & 79.56 & 85.11 & 83.69 & 90.63 & 77.28 & 43.00 \\
UI-TARS-1.5-7B  & Open   & 9.09 & 0.00 & 26.67 & 98.53 & 73.97 & 92.91 & 37.59 & 95.20 & 65.34 & 28.00 \\
GUI-owl-7B      & Open   & 0.00 & 0.00 & 20.00 & 96.32 & 81.32 & 58.87 & 56.74 & 86.65 & 74.36 & 36.50 \\

\midrule
\textit{Mobile-Aptus} & Open & 55.56 & 85.71 & 66.67 & \textbf{99.27} & \textbf{89.15} & \textbf{100.0} & 90.65 & \textbf{98.24} & \textbf{88.62} & \textbf{68.32} \\
\bottomrule
\end{tabular}

\end{table*}

\begin{table*}[h]
\centering
\renewcommand{\arraystretch}{1.2}
\setlength{\tabcolsep}{3pt}
\caption{Model Performance Comparison On AITZ Benchmark}
\label{model_performance_on_aitz_benchmark}

\begin{tabular}{lccccccccccccc}
\toprule
\multirow{2}{*}{\textbf{Models}} &
\multirow{2}{*}{\textbf{Source}} &
\multirow{2}{*}{\textbf{SCROLL}} &
\multirow{2}{*}{\textbf{PRESS}} &
\multirow{2}{*}{\textbf{STOP}} &
\multicolumn{2}{c}{\textbf{CLICK}} &
\multicolumn{2}{c}{\textbf{TYPE}} &
\multicolumn{2}{c}{\textbf{TOTAL}} &
\multirow{2}{*}{\textbf{TSR}} \\
\cmidrule(lr){6-7} \cmidrule(lr){8-9} \cmidrule(lr){10-11}
& & & & &
\textbf{Type (\%)} $\uparrow$ &
\textbf{SR (\%)} $\uparrow$ &
\textbf{Type (\%)} $\uparrow$ &
\textbf{SR (\%)} $\uparrow$ &
\textbf{Type (\%)} $\uparrow$ &
\textbf{SR (\%)} $\uparrow$ & \\
\midrule

GPT-4o          & Closed & 24.17 & 23.84 & 0.00 & 63.80 & 27.71 & 35.20 & 16.00 & 58.32 & 22.69 & 0.00 \\
GLM-4V-Plus     & Closed & 11.65 & 7.28  & 0.00 & 79.15 & 27.65 & 43.80 & 20.40 & 68.95 & 20.92 & 0.00 \\
Qwen-VL-MAX     & Closed & 7.89  & 13.04 & 10.20 & / & 72.30 & / & 34.04 & / & 52.41 & / \\

\midrule

Auto-UI         & Open   & 74.88 & 49.09 & 60.12 & 44.37 & 12.72 & 73.00 & 67.80 & 73.79 & 34.46 & / \\
Qwen2-VL-7B     & Open   & 18.64 & 21.19 & 0.00  & 71.05 & 32.89 & 82.80 & 45.00 & 66.28 & 28.25 & 0.00 \\
OS-Atlas-Pro-7B & Open   & 27.40 & 0.66  & 5.16  & 93.31 & 34.87 & 85.20 & 27.40 & 85.20 & 33.66 & 0.00 \\
CogAgent        & Open   & 56.41 & 48.30 & 4.76  & 79.90 & 51.50 & 67.40 & 34.00 & 65.86 & 44.52 & / \\
Qwen3-VL-8B     & Open   & 2.00  & 53.26 & 37.90 & 74.82 & 48.21 & 59.00 & 52.40 & 62.27 & 42.09 & 1.38 \\
UI-TARS-1.5-7B  & Open   & 0.67  & 36.81 & 31.94 & 94.33 & 50.80 & 95.40 & 23.60 & 82.64 & 38.41 & 0.59 \\
GUI-owl-7B      & Open   & 1.17  & 54.83 & 18.85 & 75.22 & 43.60 & 58.20 & 50.20 & 60.81 & 37.18 & 0.20 \\

\midrule
\textit{Mobile-Aptus} & Open & \textbf{88.12} & \textbf{66.44} & \textbf{88.02} &
\textbf{96.51} & \textbf{87.02} &
\textbf{98.60} & \textbf{80.16} &
\textbf{98.36} & \textbf{85.82} & \textbf{20.95} \\
\bottomrule
\end{tabular}
\end{table*}

\begin{table*}[!htbp]
\centering
\renewcommand{\arraystretch}{1.2}
\setlength{\tabcolsep}{3pt}
\caption{Model Performance Comparison On Meta-GUI Benchmark}
\label{model_performance_on_metagui_benchmark}

\begin{tabular}{lccccccccccccc}
\toprule
\multirow{2}{*}{\textbf{Models}} &
\multirow{2}{*}{\textbf{Source}} &
\multirow{2}{*}{\textbf{SCROLL}} &
\multirow{2}{*}{\textbf{PRESS}} &
\multirow{2}{*}{\textbf{STOP}} &
\multicolumn{2}{c}{\textbf{CLICK}} &
\multicolumn{2}{c}{\textbf{TYPE}} &
\multicolumn{2}{c}{\textbf{TOTAL}} &
\multirow{2}{*}{\textbf{TSR}} \\
\cmidrule(lr){6-7} \cmidrule(lr){8-9} \cmidrule(lr){10-11}
& & & & &
\textbf{Type (\%)} $\uparrow$ &
\textbf{SR (\%)} $\uparrow$ &
\textbf{Type (\%)} $\uparrow$ &
\textbf{SR (\%)} $\uparrow$ &
\textbf{Type (\%)} $\uparrow$ &
\textbf{SR (\%)} $\uparrow$ & \\
\midrule

GPT-4o          & Closed & 33.97 & 25.00 & 12.79 & 94.12 & 42.30 & 66.47 & 28.14 & 69.85 & 32.72 & 6.67 \\
GLM-4V-Plus     & Closed & 0.00  & 0.00  & 1.06  & 95.45 & 26.53 & 38.01 & 22.81 & 65.05 & 17.19 & 1.67 \\
Qwen-VL-MAX     & Closed & 14.74 & 40.91 & 1.91  & 70.87 & 37.85 & 74.85 & 45.03 & 54.74 & 27.86 & 15.42 \\

\midrule

Auto-UI         & Open   & 60.90 & 0.00  & 0.00  & 26.90 & 2.69  & 0.00  & 0.00  & 20.02 & 6.45  & 0.00 \\
Qwen2-VL-7B     & Open   & 0.00  & 0.00  & 0.43  & 51.54 & 2.33  & 38.01 & 18.17 & 35.90 & 3.04  & 0.21 \\
OS-Atlas-Pro-7B & Open   & 16.03 & 0.00  & 0.00  & 94.53 & 37.29 & 60.23 & 15.20 & 66.09 & 23.59 & 0.42 \\
Qwen3-VL-8B     & Open   & 13.29 & 6.67  & 0.00  & 88.77 & 53.03 & 62.57 & 54.97 & 77.54 & 47.28 & 9.70 \\
UI-TARS-1.5-7B  & Open   & 12.66 & 6.67  & 0.00  & 92.85 & 36.47 & 33.33 & 27.49 & 76.45 & 31.35 & 5.66 \\
GUI-owl-7B      & Open   & 0.63  & 6.67  & 0.00  & 87.60 & 46.39 & 63.16 & 47.95 & 77.77 & 41.38 & 9.97 \\

\midrule
\textit{Mobile-Aptus} & Open &
\textbf{98.08} & \textbf{97.73} & \textbf{94.73} &
\textbf{99.35} & \textbf{96.94} &
\textbf{99.42} & \textbf{95.91} &
\textbf{98.08} & \textbf{96.41} & \textbf{88.33} \\
\bottomrule
\end{tabular}
\end{table*}

\begin{table*}[!htbp]
\centering
\renewcommand{\arraystretch}{1.2}
\setlength{\tabcolsep}{3pt}
\caption{Model Performance Comparison On AndroidControl Benchmark}
\label{model_performance_on_androidcontrol_benchmark}

\begin{tabular}{lccccccccccc}
\toprule
\multirow{2}{*}{\textbf{Models}} &
\multirow{2}{*}{\textbf{Source}} &
\multirow{2}{*}{\textbf{SCROLL}} &
\multirow{2}{*}{\textbf{PRESS}} &
\multirow{2}{*}{\textbf{STOP}} &
\multicolumn{2}{c}{\textbf{CLICK}} &
\multicolumn{2}{c}{\textbf{TYPE}} &
\multicolumn{2}{c}{\textbf{TOTAL}} &
\multirow{2}{*}{\textbf{TSR}} \\
\cmidrule(lr){6-7} \cmidrule(lr){8-9} \cmidrule(lr){10-11}
& & & & &
\textbf{Type (\%)} $\uparrow$ &
\textbf{SR (\%)} $\uparrow$ &
\textbf{Type (\%)} $\uparrow$ &
\textbf{SR (\%)} $\uparrow$ &
\textbf{Type (\%)} $\uparrow$ &
\textbf{SR (\%)} $\uparrow$ & \\
\midrule

GPT-4o          & Closed & 18.21 & 0.00 & 13.16 & 36.12 & 21.31 & 15.63 & 34.25 & 61.98 & 27.62 & 5.06 \\
GLM-4V-Plus     & Closed & 1.66  & 0.00 & 0.71  & 24.09 & 21.04 & 36.30 & 7.06  & 61.95 & 19.20 & 2.92 \\
Qwen-VL-MAX     & Closed & 1.92  & 0.00 & 12.37 & 42.28 & 43.99 & 16.42 & 11.11 & 61.07 & 31.28 & 6.16 \\

\midrule

Auto-UI         & Open   & 23.24 & 0.00 & 0.00  & 1.32  & 0.00  & 0.00  & 0.00  & 16.28 & 4.12  & 0.00 \\
Qwen2-VL-7B     & Open   & 0.08  & 0.00 & 1.67  & 10.21 & 25.97 & 25.21 & 0.00  & 40.24 & 9.70  & 0.48 \\
OS-Atlas-Pro-7B & Open   & 6.52  & 0.00 & 53.62 & 43.56 & 44.30 & 3.29  & 0.58  & 69.22 & 34.28 & 2.46 \\
Qwen3-VL-8B     & Open   & 28.16 & 49.85 & 35.27 & 89.04 & \textbf{61.41} & 83.39 & 73.58 & 73.80 & 54.20 & 17.30 \\
UI-TARS-1.5-7B  & Open   & 0.20  & 27.11 & 26.18 & \textbf{92.69} & 54.73 & 85.44 & 47.63 & 83.78 & 43.22 & 12.71 \\
GUI-owl-7B      & Open   & 42.94 & 57.73 & 69.63 & 84.43 & 60.94 & 85.76 & 78.32 & 75.61 & 57.84 & 20.05 \\

\midrule
\textit{Mobile-Aptus} & Open &
\textbf{50.70} & \textbf{88.89} & \textbf{81.27} &
81.82 & 54.11 &
\textbf{89.46} & \textbf{85.13} &
\textbf{91.64} & \textbf{75.81} & \textbf{40.12} \\
\bottomrule
\end{tabular}
\end{table*}

\subsection{Main Results}

Tables~\ref{model_performance_on_kairos_dataset}, \ref{model_performance_on_aitz_benchmark}, \ref{model_performance_on_metagui_benchmark} and~\ref{model_performance_on_androidcontrol_benchmark} present the results of various baselines and Mobile-Aptus on the OS-Kairos dataset, AITZ benchmark, Meta-GUI benchmark, and AndroidControl benchmark. 
Table~\ref{comparison} shows a comparison between OS-Kairos and Mobile-Aptus on these datasets and benchmarks in terms of task completion performance, adaptive interaction accuracy, and number of invention steps. 
Based on these results, we have the following key findings:

(i) \textbf{Mobile-Aptus demonstrates strong task completion performance, achieving state-of-the-art results across various datasets and benchmarks.} As shown in Tables~\ref{model_performance_on_kairos_dataset}, \ref{model_performance_on_aitz_benchmark}, \ref{model_performance_on_metagui_benchmark}, and~\ref{model_performance_on_androidcontrol_benchmark}, Mobile-Aptus outperforms the best-performing baselines in both SR and TSR metrics. 
Moreover, compared to zero-shot models, it achieves improvements in SR ranging from 27.26\% to 72.82\%, and in TSR from 20.95\% to 87.91\% across different datasets and benchmarks.
(Relative to OS-Atlas-Pro-7B, note that we have not conducted any training on the action-generation capability of OS-Atlas-Pro-7B.)
These results show that adaptive interaction for mobile-using agents leads to significant improvements in task completion performance compared to fully automated mobile-using agents.

(ii) \textbf{Compared to OS-Kairos, Mobile-Aptus has more accurate timing for intervention.} As shown in Table~\ref{comparison}, on the OS-Kairos dataset, Mobile-Aptus improves IP from 70.75\% to 84.83\%, effectively reducing over-soliciting interventions to about half of the original rate, while also improving the HSR. 
As a result, the distinction between necessary and unnecessary human intervention steps in these benchmarks is inherently ambiguous. Nonetheless, Mobile-Aptus still achieves improvements in both IP and HSR across the three benchmarks. Therefore, these benchmark experiments serve as supplementary validation of the effectiveness of Mobile-Aptus.
In conclusion, the results indicate that Mobile-Aptus achieves more precise timing for intervention.


(iii) \textbf{The lower the performance of the probed mobile-using agent on a given dataset or benchmark, the greater the improvement achieved by Mobile-Aptus over the probed mobile-using agent.} As shown in Tables~\ref{model_performance_on_kairos_dataset}, \ref{model_performance_on_aitz_benchmark}, \ref{model_performance_on_metagui_benchmark}, and~\ref{model_performance_on_androidcontrol_benchmark}, under the zero-shot setting, the smaller the SR of the probed mobile-using agent, the larger the SR gain brought by Mobile-Aptus. This is because, on the one hand, datasets or benchmarks with lower SR provide more opportunities for intervention-based learning, enabling Mobile-Aptus to better understand the task context. On the other hand, when the probed mobile-using agent makes more incorrect actions, there are more issues that can be addressed through intervention, resulting in a larger performance gain.

In summary, Mobile-Aptus  eases the challenge of over-execution through state-of-the-art task completion performance, and mitigates the challenge of over-soliciting by making more accurate decisions on when to intervene. 
Notably, both improvements are achieved with only a small cost in terms of intervention steps.

\begin{table*}[t]
    \centering
    \caption{Comparison of Metrics between OS-Kairos and Mobile-Aptus}
    \label{comparison}
    \renewcommand{\arraystretch}{1.2}
    \setlength{\tabcolsep}{6pt}

    \begin{tabular}{lcccccccc}
        \toprule
        \multirow{2}{*}{Benchmark} &
        \multicolumn{2}{c}{\textbf{IP}(\%)$\uparrow$} &
        \multicolumn{2}{c}{\textbf{HSR}(\%)$\uparrow$} &
        \multicolumn{2}{c}{\textbf{AIF}$\downarrow$} &
        \multicolumn{2}{c}{\textbf{SR}(\%)$\uparrow$} \\
        \cmidrule(lr){2-3} \cmidrule(lr){4-5}
        \cmidrule(lr){6-7} \cmidrule(lr){8-9}
        & OS-Kairos & Mobile-Aptus
        & OS-Kairos & Mobile-Aptus
        & OS-Kairos & Mobile-Aptus
        & OS-Kairos & Mobile-Aptus \\
        \midrule
        OS-Kairos        & 70.75 & 84.83 & 86.87 & 88.62 & 1.59 & 1.06 & 95.90 & 88.62 \\
        AITZ             & 71.64 & 73.27 & 72.20 & 72.54 & 5.78 & 5.42 & 87.54 & 85.82 \\
        Meta-GUI         & 93.18 & 93.25 & 91.73 & 91.73 & 4.28 & 4.28 & 96.36 & 96.41 \\
        AndroidControl   & 77.77 & 77.96 & 74.89 & 72.84 & 2.93 & 2.75 & 79.78 & 75.81 \\
        \bottomrule
    \end{tabular}
\end{table*}

\subsection{Analysis} \label{sec:analysis}
We first analyze the performance of Mobile-Aptus in dynamic real-world evaluation. We then demonstrate the generality of Mobile-Aptus through model scaling experiments, and validate the necessity of selecting the top-k most similar samples via a scaling experiment on the value of k. Next, we illustrate the superiority of adaptive interaction through an interactive sensitivity analysis. Finally, we explain why Mobile-Aptus outperforms approaches that directly determine the time for intervention through prompting, and further explore whether human intervention can be effectively replaced by an automated agent or agent system.

\subsubsection{\textbf{Dynamic Realworld Evaluation}}
Our experiments on existing datasets and benchmarks fall under static evaluation. To demonstrate the effectiveness of our work, we further conducted dynamic real-world evaluation experiments on real Android devices using the Android Debug Bridge (ADB).

We randomly selected 50 instructions from the test set of the OS-Kairos dataset for testing on real devices and manually evaluated whether each episode was successfully executed. Additionally, if the execution exceeds 10 steps, the trajectory is automatically terminated, and the next instruction begins. The experimental results are shown in Table~\ref{Dynamic Realworld Evaluation Experiment}.

    
\begin{figure*}[t]
    \centering
    \includegraphics[width=1\linewidth]{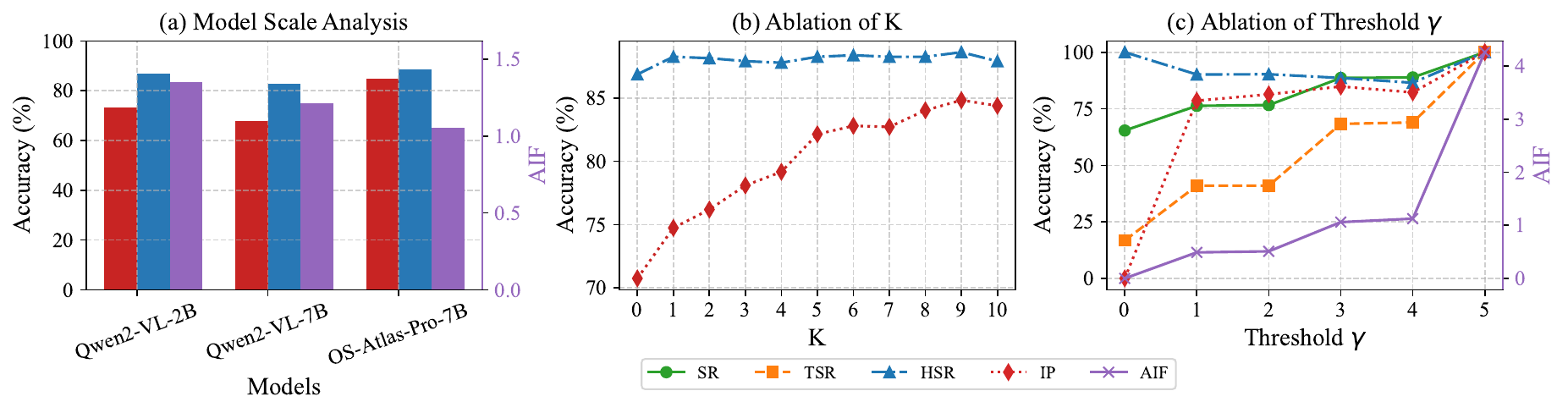}
    \vspace{-0.4cm}
    \caption{(a) Results of the model scaling experiment. The results demonstrate the rationality of the probed mobile-using agent selection and the strong generalization ability of Mobile-Aptus. (b) Results of the k scale experiment results. The results show that the necessity of selecting the top-\(k\) most similar samples and further confirm the effectiveness of the confidence bias correction strategy based on semantic similarity retrieval and DPO. (c) Results of the interactive sensitivity experiment. The results demonstrate the flexibility of adaptive interaction mobile-using 5agents.}
    \vspace{-0.4cm}
    \label{analyze}
\end{figure*}

\begin{table}[t]
    \centering
    \renewcommand{\arraystretch}{1.1}
    \caption{Dynamic Realworld Evaluation Experiment}
    \label{Dynamic Realworld Evaluation Experiment}
    \resizebox{\linewidth}{!}{
    \begin{tabular}{lcccc}
    
    \toprule 
    \textbf{Models}  & \textbf{Execution Steps} & \textbf{RE (\%)}$\uparrow$ & \textbf{TSR (\%)}$\uparrow$ & \textbf{AIF}$\downarrow$ \\ \midrule
    GPT-4o & 302 & 75.83 & 36.00 & /\\
    Qwen2-VL-7B  & 397 & 57.68 & 4.00 & /\\
    OS-Atlas-Pro-7B & 359 & 63.79 & 26.00& / \\ \midrule
    OS-Kairos  & 265 & 86.42 & 70.00 & 0.96\\
    Mobile-Aptus  &246  &93.09 &62.00	& 0.64 \\ \bottomrule
    \end{tabular}}
    \vspace{-0.4cm}
\end{table}
Mobile-Aptus is mainly compared against two types of agents: automated mobile-using agents and interactive agents.
For automated mobile-using agents, we compare TSR to measure task completion ability, hence the "36\% improvement."
For interactive agents, TSR alone should not be the only metric. If interactive agents require human intervention at every step, their TSR would be 100\%. Therefore, RE and AIF are more informative metrics. Mobile-Aptus achieves a 6.67\% improvement in RE over OS-Kairos and reduces AIF by 0.32.

\subsubsection{\textbf{General Effectiveness across Scales}}We will analyze the impact of model scale and the scale of the top-k similarity matches in constructing $\mathcal{D}_{\text{DPO}}$ on the effectiveness of our work.


\textbf{Model scale.} The probed mobile-using agent in this work adopts OS-Atlas-Pro-7B. 
To investigate the impact of model scale on our framework, we conduct a fair comparison by training Mobile-Aptus using Qwen2-VL-2B, Qwen2-VL-7B, and OS-Atlas-Pro-7B within the same framework. 
As shown in Figure~\ref{analyze}, OS-Atlas-Pro-7B achieves the highest IP and HSR while maintaining the lowest AIF, indicating that as the probed mobile-using agent, OS-Atlas-Pro-7B enables Mobile-Aptus to request human intervention with the highest accuracy.

\label{k_scale}
\textbf{K scale.} In the DPO stage, we construct the dataset \(\mathcal{D}_{\text{DPO}}\) by selecting the top-\(k\) most similar samples for incorrectly classified confidence score samples to form positive and negative pairs. Figure~\ref{analyze} illustrates the HSR and IP of Mobile-Aptus as \(k\) varies from 0 to 10. 
As \(k\) increases, both HSR and IP exhibit an upward trend, with IP increasing significantly. 
This indicates that our DPO algorithm effectively enhances the accuracy of human intervention for Mobile-Aptus. Eventually, the growth of IP stabilizes and reaches its peak. 
This demonstrates the necessity of selecting the top-\(k\) most similar samples and further confirms the effectiveness of the confidence bias correction strategy based on semantic similarity retrieval and DPO.


\subsubsection{\textbf{Interactive Sensitivity Analysis}}

Mobile-Aptus possesses adaptive interaction capabilities. Users can adjust the user-set intervention threshold \(\gamma\) to control the sensitivity of human intervention requests. 
When the confidence score provided by Mobile-Aptus is \(\leq \gamma\), it is considered necessary to request human intervention. Since the output score of Mobile-Aptus ranges from 1 to 5, setting \(\gamma = 0\) means no human intervention is requested at any step, while \(\gamma = 5\) results in a request at every step. 
A larger \(\gamma\) increases the likelihood of requesting human intervention. We evaluated Interactive Sensitivity by analyzing Mobile-Aptus’s performance under different \(\gamma\) values. 
Figure~\ref{analyze} illustrates the variations in different metrics under various values of \(\gamma\). 
The results indicate that Mobile-Aptus can flexibly adjust its Interactive Sensitivity based on the user's chosen \(\gamma\), allowing adaptation to different scenarios. 
For instance, when task completion is prioritized, a larger \(\gamma\) can be selected, whereas a smaller \(\gamma\) is preferable when minimizing interruptions is the primary concern.

\subsubsection{\textbf{Comparing with prompt-based interaction}}
Adaptive interaction agents can be developed in two ways: (1) using data-driven domain adaptation methods such as Mobile-Aptus, or (2) prompting the agent to output a confidence score directly. We compare these approaches on the OS-Kairos dataset, as shown in Table~\ref{HCI intervention}. The performance of the prompt-based method for deciding whether to request human intervention shows no significant improvement over the zero-shot action prediction results (Table~\ref{model_performance_on_kairos_dataset}), with the best-performing model, GPT-4o, achieving only a 46.58\% TSR. In contrast, Mobile-Aptus achieves a TSR of 68.32\%, outperforming the best baseline by 21.74\%. These results suggest that prompting agents to output confidence scores is ineffective due to their lack of self-awareness regarding capability boundaries, whereas the Mobile-Aptus approach proves to be effective.

\begin{table}[t]
    \centering
    \LARGE
    \renewcommand{\arraystretch}{1.2}
    \caption{Analysis of interactive paradigms vs. prompt-based baseline in OS-Kairos dataset}
    \resizebox{\linewidth}{!}{
    \begin{tabular}{lccccc}
    \toprule 
    \textbf{Models} & \textbf{Interactive} & \textbf{Type (\%)}$\uparrow$ & \textbf{SR (\%)}$\uparrow$ & \textbf{TSR (\%)}$\uparrow$ &\\ \midrule
    GPT-4o & Prompt &88.80 &79.25 &46.58 \\
    GLM-4V-Plus & Prompt &88.34 &79.03 &47.83  \\ 
    Qwen2-VL-7B & Prompt &76.42 &38.44 &25.47   \\
    OS-Atlas-Pro-7B & Prompt &95.67 &59.02 &9.94
    \\ \midrule
    \textit{Mobile-Aptus} & FT+DPO & \textbf{98.24} & \textbf{88.62} &\textbf{68.32} \\ \bottomrule
    \end{tabular}}
    \label{HCI intervention}
    \vspace{-0.3cm}
\end{table}

\subsubsection{\textbf{Edge-Cloud Collaboration for Replacing Human Intervention}}

In Mobile-Aptus, human intervention can be replaced by a multi-agent system when task interruption is undesirable. 
As shown in Table~\ref{gpt}, we substitute the human with a multi-agent system constructed using GPT-4o. This achieves a 32.91\% improvement in TSR without human intervention. 
While such cloud-based systems may introduce latency and privacy concerns, they offer an effective alternative for adaptive interaction when human participation is not feasible.
Since only a few steps require human intervention, the execution latency of this architecture is also acceptable.
A cost-effectiveness analysis shows that an average latency increase of 0.56 seconds brings a 32.91\% improvement in TSR.

\begin{table}[t]
    \centering
    \LARGE
    \renewcommand{\arraystretch}{1.2}
    \caption{Analysis of human intervention vs. multi-agent system intervention in the OS-Kairos dataset}
    \resizebox{\linewidth}{!}{
    \begin{tabular}{lccccc}
    \toprule 
    \textbf{Models} & \textbf{Interactive} & \textbf{Type (\%)}$\uparrow$ & \textbf{SR (\%)}$\uparrow$ & \textbf{TSR (\%)}$\uparrow$ &\\ \midrule
    OS-Atlas-Pro-7B & / &95.90 &61.36 &14.29 \\
    \textit{Mobile-Aptus}  &Multi-agent &98.12$_{\textcolor[RGB]{178,34,34}{2.22\uparrow}}$  & 82.30$_{\textcolor[RGB]{178,34,34}{20.94\uparrow}}$ &47.20$_{\textcolor[RGB]{178,34,34}{32.91\uparrow}}$  \\ 
    \textit{Mobile-Aptus}  &Human &98.24$_{\textcolor[RGB]{178,34,34}{2.34\uparrow}}$  &88.62$_{\textcolor[RGB]{178,34,34}{27.26\uparrow}}$ &68.32$_{\textcolor[RGB]{178,34,34}{54.03\uparrow}}$  \\ 
    \bottomrule
    \end{tabular}}
    \label{gpt}
    \vspace{-0.3cm}
\end{table}

\subsubsection{\textbf{Visual analytics}}

\begin{figure}[t]
    \centering
    \includegraphics[width=1\linewidth]{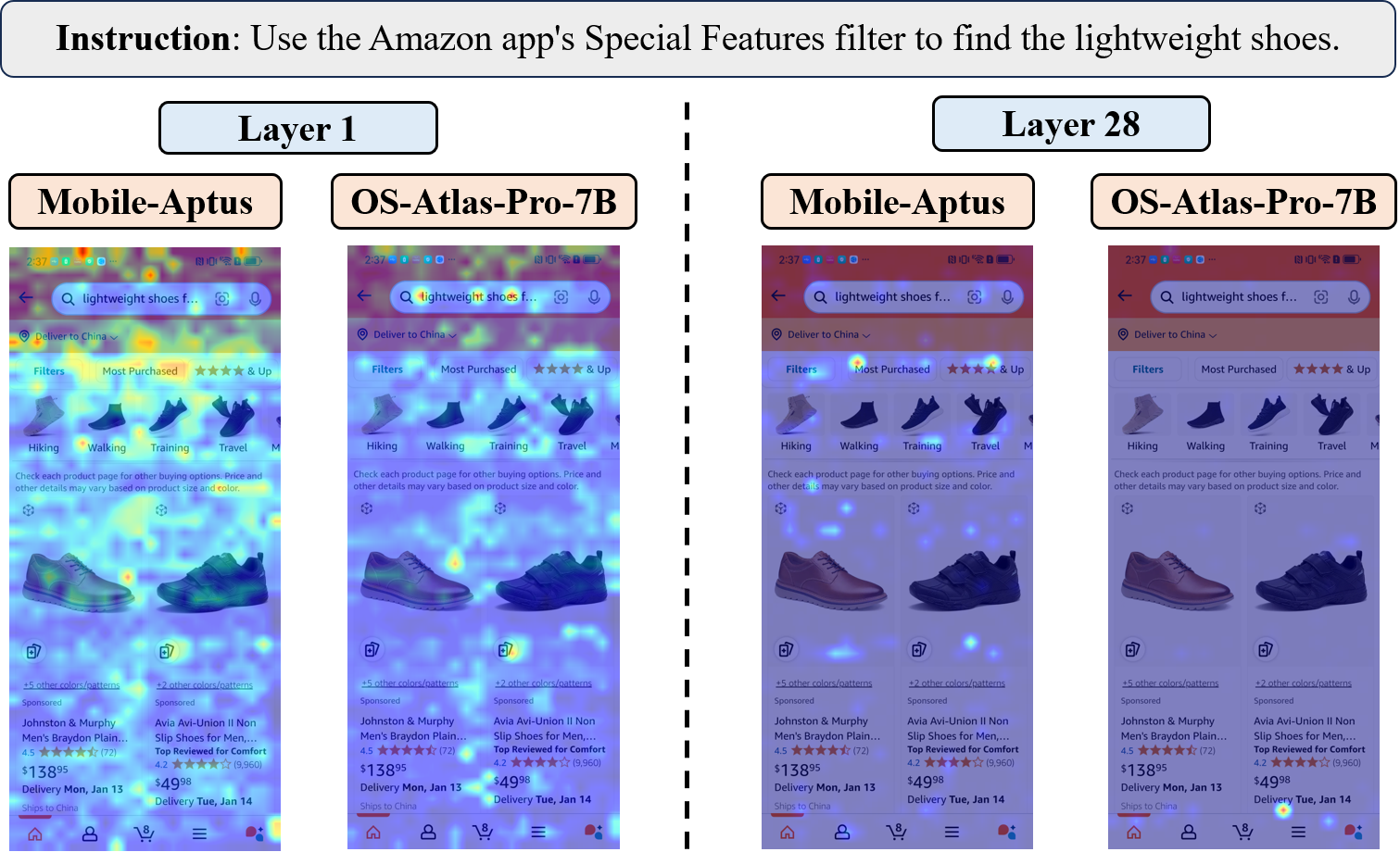}
    \vspace{-0.4cm}
    \caption{Comparison of attention heatmaps between Mobile-Aptus and OS-Atlas-Pro-7B. Mobile-Aptus exhibits a globally distributed attention pattern and more accurate visual component attention related to the instruction.}
    \vspace{-0.4cm}
    \label{vision}
\end{figure}

As shown in Figure~\ref{vision}, we visualize the attention heatmaps~\cite{xu2025attention} of Mobile-Aptus and OS-Atlas-Pro-7B to explain why Mobile-Aptus is better at identifying situations that require human intervention. 
We observe that in lower layers (e.g., Layer 1), Mobile-Aptus exhibits a more globally distributed attention pattern compared to OS-Atlas-Pro-7B, suggesting a more holistic understanding of the screenshot. 
This may help Mobile-Aptus avoid reliance on shortcuts that result from overfitting. 
Furthermore, in higher layers (e.g., the final Layer 28), the attention in OS-Atlas-Pro-7B fails focus on the “Filter” button, whereas Mobile-Aptus accurately attends to this region. This enables Mobile-Aptus to correctly infer that the current step, which omits clicking the filter button, should be assigned a lower confidence score.
In summary, the relationship between attention patterns and confidence scores can be summarized in two aspects: i) Introducing the ability for the agent to output confidence scores leads the agent’s attention patterns to become more globalized. ii) Outputting confidence scores allows Mobile-Aptus to attend, in deeper layers, to components that OS-Atlas overlooks.

\subsubsection{\textbf{Time cost analysis}}

To evaluate the computational overhead of the Mobile-Aptus training pipeline, we provide a detailed breakdown of the time consumed by each stage when constructing the model using the OS-Kairos dataset.
\begin{table}[t]
\centering
\caption{Efficiency analysis of the model alignment process. We report the training time breakdown for Mobile-Aptus and compare its inference latency with the backbone model OS-Atlas-pro-7B.}
\label{tab:efficiency_analysis}
\small
\begin{tabular}{llcc}
\toprule
\textbf{Category} & \textbf{Stage / Model} & \textbf{Time / Latency} & \textbf{Ratio / $\Delta$} \\ \midrule
\multirow{4}{*}{Training} & SFT & 1:46:02 & 52.4\% \\
 & Top-$k$ Matching & 0:06:37 & 3.3\% \\
 & DPO & 1:29:43 & 44.3\% \\ \cmidrule{2-4} 
 & \textbf{Total Training} & \textbf{3:22:22} & \textbf{100.0\%} \\ \midrule
\multirow{2}{*}{Inference} & OS-Atlas-pro-7B & 1.02 s & -- \\
 & \textbf{Mobile-Aptus} & \textbf{1.11 s} & \textbf{+0.09 s} \\ \bottomrule
\end{tabular}
\end{table}
As illustrated in Table~\ref{tab:efficiency_analysis}, compared to the parameter update phases (SFT and DPO), the Top-k matching stage requires only simple input encoding and vector matching via the encoder. This results in a significantly lower computational burden, thereby demonstrating the high efficiency of our proposed confidence bias correction stage.
Notably, the finalized Mobile-Aptus maintains competitive inference speed with an average latency of 1.11s. Compared to the backbone OS-Atlas-pro-7B (1.02s), the marginal increase in latency (+0.09s) is negligible for real-time applications, confirming that our alignment methodology enhances model performance without compromising deployment feasibility.

\subsubsection{\textbf{Ablation Study}}
To investigate the individual contributions of each stage in our pipeline, we conducted an ablation study on the components used to construct Mobile-Aptus from the OS-Kairos dataset.

\begin{table}[t]
\centering
\caption{Ablation study of Mobile-Aptus. We compare the performance of different training stages and components. The best results are highlighted in \textbf{bold}.}
\label{tab:ablation_study}
\begin{tabular}{lcccc}
\toprule
Method & IP (\%)$\uparrow$ & HSR (\%)$\uparrow$ & AIF$\downarrow$ & SR (\%)$\uparrow$ \\
\midrule
Base  & -   & -    & -  & 61.36 \\
SFT  & 70.75 & 86.87 & 1.59 & \textbf{95.90} \\
SFT + DPO w/o retrieval & 74.89 & 87.44 & 1.41 & 91.42 \\
Mobile-Aptus & \textbf{84.83} & \textbf{88.62} & \textbf{1.06} & 88.62 \\
\bottomrule
\end{tabular}
\end{table}

As summarized in Table~\ref{tab:ablation_study}, while SFT alone equips the base model with the basic capability to output confidence scores, the subsequent addition of the retrieval mechanism and DPO yields consistent performance gains across the IP, HSR, and AIF metrics. These results validate the design of the Mobile-Aptus construction pipeline.

It is worth noting that while SR is a standard metric for evaluating a mobile agent's task completion, it does not fully reflect the performance of an interactive agent. 
The combination of low IP and high SR observed in the SFT model indicates that the agent frequently requests human intervention, even in scenarios where it is unnecessary. 
In an extreme case, an agent that triggers human assistance at every step would theoretically achieve a 100\% SR; however, such a model fails to fulfill the role of an efficient autonomous assistant. Thus, the improvement in IP and AIF demonstrated by Mobile-Aptus highlights its superior ability to balance autonomy with necessary human collaboration.

\subsubsection{\textbf{Trade-off between intervention and per-step accuracy}}
For interactive agents, it is necessary to consider both the intervention proportion and the per-step accuracy. As shown in Figure~\ref{analyze}, interactive agents that only hack the per-step accuracy (with the threshold set to 5) tend to require human intervention at every step. Although they achieve 100\% SR, their HSR is poor. 
Therefore, while per-step accuracy should be the primary concern for fully autonomous agents, it is more reasonable to focus on IP and HSR for interactive agents.

\subsubsection{\textbf{Out-of-distribution experiment}}
We evaluated OS-Kairos and Mobile-Aptus, both trained on the OS-Kairos dataset, on the AITZ benchmark. The experimental results show that the AIF of OS-Kairos is 6.42, and the AIF of Mobile-Aptus is 5.18. 
The experimental results further demonstrate that when faced with new UI environments, specifically benchmarks that were not included during training with confidence scores, OS-Kairos exhibits an extremely high AIF due to overconfidence, whereas Mobile-Aptus mitigates this issue.

\section{Conclusion}\label{sec6}
This work proposes a universal confidence integration framework that enables fully autonomous mobile-using agents to generate confidence scores for their actions, allowing them to request human intervention at appropriate moments. Experiments on our dataset and three popular mobile-using agents benchmarks demonstrate that Mobile-Aptus effectively addresses the over-execution problem in fully autonomous agents. Furthermore, Mobile-Aptus mitigates the over-soliciting issue, where agents request human intervention unnecessarily. We demonstrate that our method effectively addresses the over-soliciting problem through a confidence bias correction strategy based on semantic similarity retrieval and DPO. Further experiments validate the adaptive interaction capability of Mobile-Aptus, allowing users to adjust intervention sensitivity autonomously. 
When the probed agent of Mobile-Aptus is replaced, the entire confidence annotation process needs to be rerun. Future work will adopt more flexible solutions to obtain confidence scores.

\bibliographystyle{ieeetr}
\bibliography{manuscript}

 \begin{IEEEbiography}[{\includegraphics[width=1in,height=1.25in,clip,keepaspectratio]{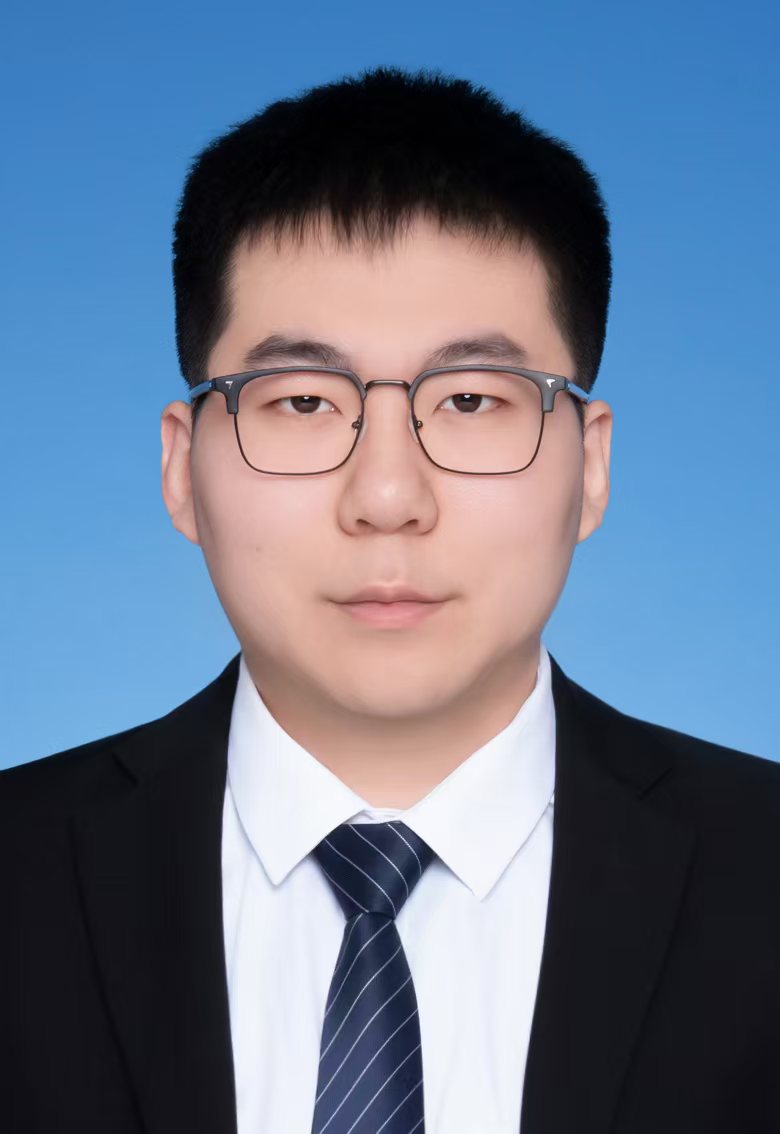}}]{Zheng Wu}
received his Bachelor's degree in information security from Shanghai Jiao Tong University, Shanghai, China, in 2025. He is working towards his M.S. degree at the AI Security Lab of Shanghai Jiao Tong University. His research interests include natural language processing, multimodal large language model, and AI agent.
\end{IEEEbiography}

\begin{IEEEbiography}[{\includegraphics[width=1in,height=1.3in,clip,keepaspectratio]{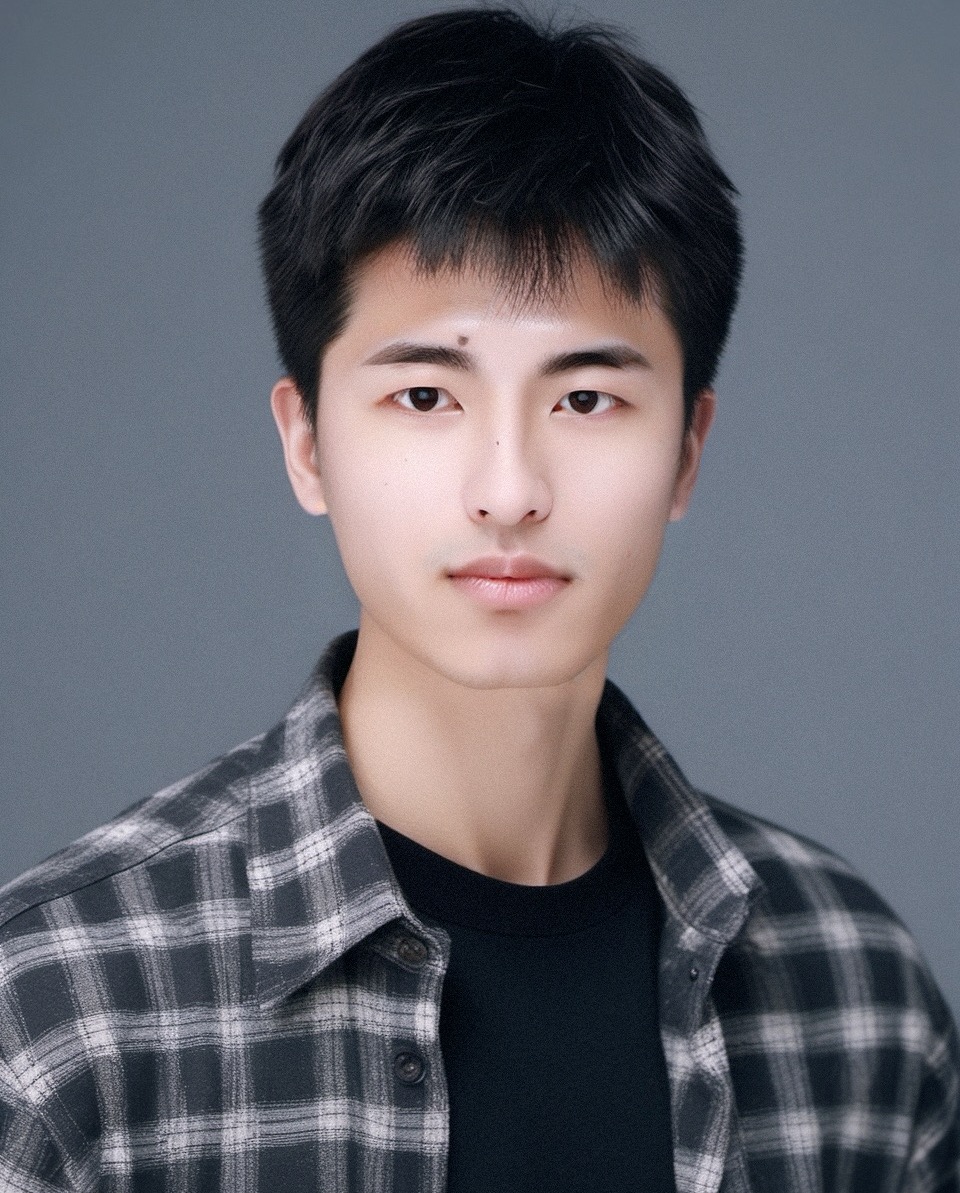}}]{Pengzhou Cheng} is an Assistant Professor in the School of Computer Engineering and Science at Shanghai University, Shanghai, China. He received his Ph.D. degree from the School of Computer Science, Shanghai Jiao Tong University, Shanghai 201100, China. His research interests include LLM reasoning, AI agents, artificial intelligence security, and cybersecurity. He has published papers in leading journals and conferences, including TNNLS, TVT, ACL, AAAI, EMNLP, and NAACL. He serves as a standing reviewer for TPAMI, TDSC, TITS, and TIFS, and has also served as a reviewer for major conferences such as NeurIPS, ACL, IJCAI, EMNLP, and COLING.\end{IEEEbiography}

\begin{IEEEbiography}[{\includegraphics[width=1in,height=1.25in,clip,keepaspectratio]{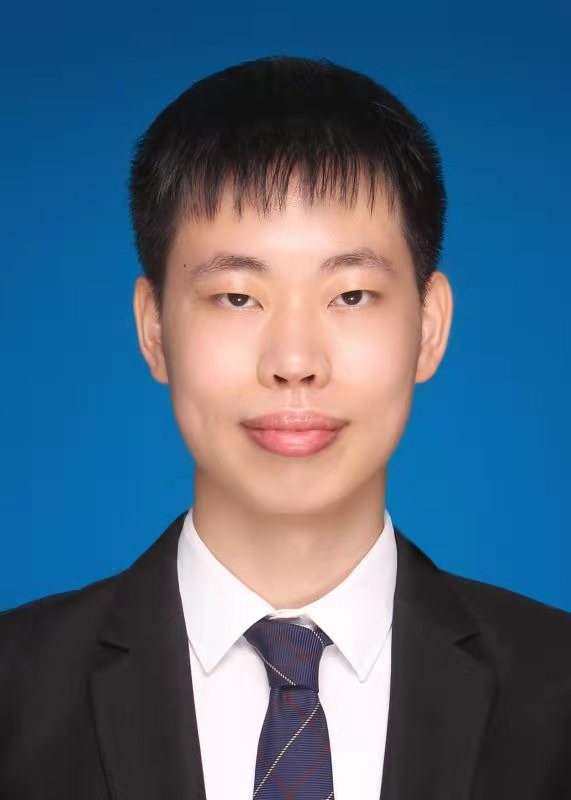}}]{Zongru Wu}
received the B.S Degree from the School of Cyber Science and Engineering, Wuhan University, Hubei, China, in 2022. He is currently persuing the Ph.D. Degree with the School of Cyber Science and Engineering, Shanghai Jiao Tong University, Shanghai, 201100, China. His primary research interests include artificial intelligence security, backdoor attack and countermeasures, cybersecurity, machine learning, and deep learning.
\end{IEEEbiography}

\begin{IEEEbiography}[{\includegraphics[width=1in,height=1.25in,clip,keepaspectratio]{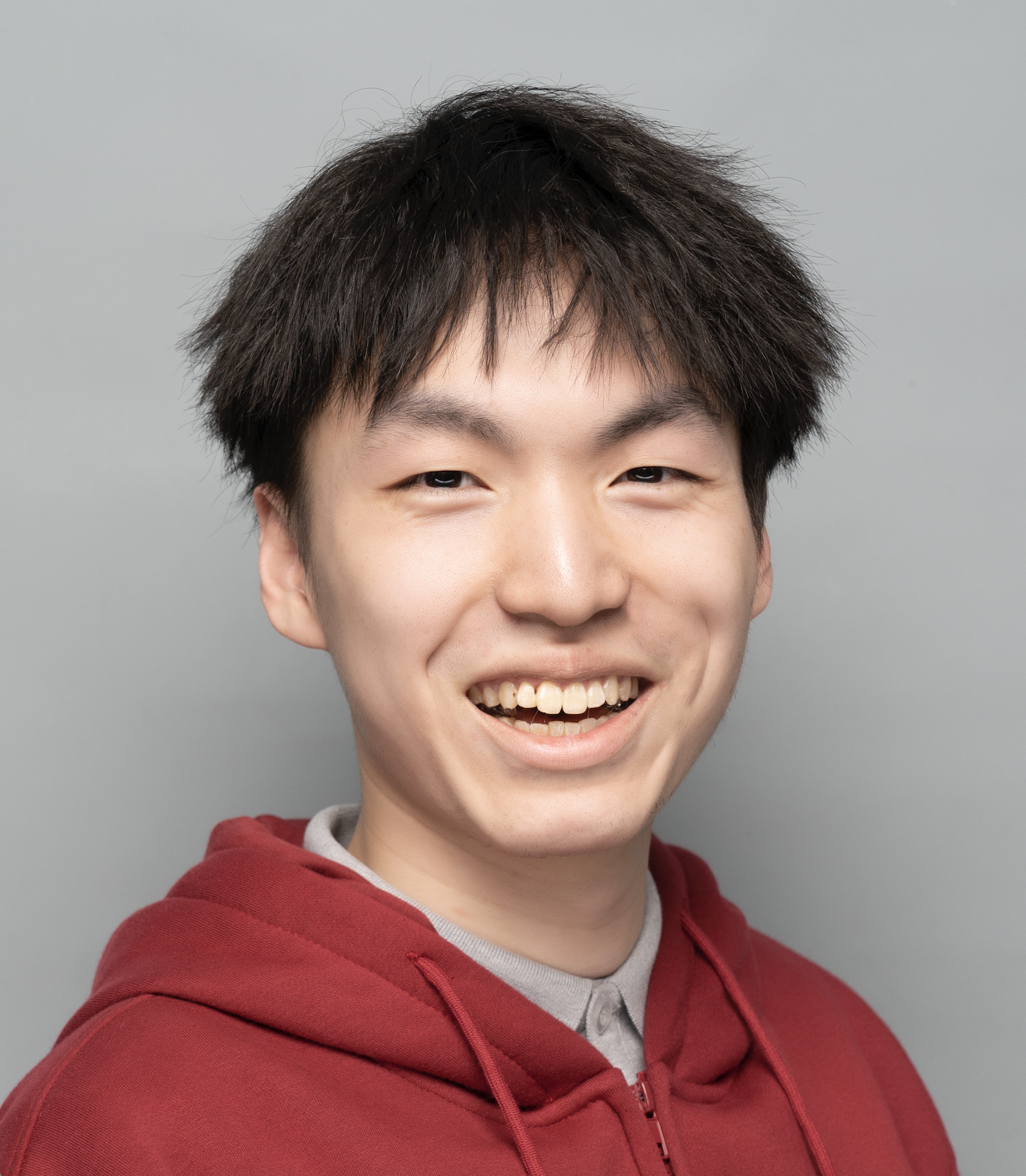}}]{Yuan Guo} is an undergraduate student majoring in computer science and technology in the Department of Computer Science and Engineering, Shanghai Jiao Tong University, Shanghai, 201100, China. His primary research interests include natural language processing, large language model and AI agent.
\end{IEEEbiography}

\begin{IEEEbiography}[{\includegraphics[width=1in,height=1.25in,clip,keepaspectratio]{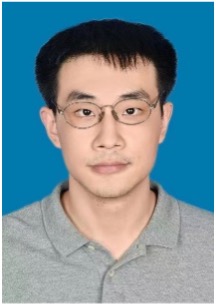}}]{Tianjie Ju} received the B.S Degree from the School of Computer Science and Software Engineering, Xidian University, Xi'an, China, in 2021. He is currently persuing the Ph.D. Degree with the School of Cyber Science and Engineering, Shanghai Jiao Tong University, Shanghai, 201100, China. His primary research interests include artificial intelligence security and interpretability, LLM-based multi-agent system, knowledge editing.
\end{IEEEbiography}

\begin{IEEEbiography}
 [{\includegraphics[width=1in, height=1.25in, clip, keepaspectratio]{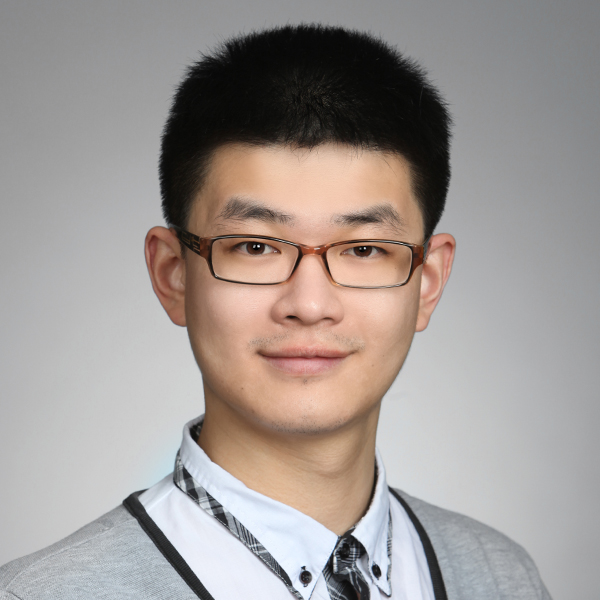}}]{Aston Zhang}
is a research scientist at Meta. His work has been recognized with the ICLR Outstanding Paper Award, the ACM Ubicomp Distinguished Paper Award, and an ACM SenSys Best Paper Award nomination. His textbook, ``Dive into Deep Learning,'' is adopted worldwide. He earned his Ph.D. in Computer Science from UIUC.
\end{IEEEbiography}

\begin{IEEEbiography}
 [{\includegraphics[width=1in, height=1.25in, clip, keepaspectratio]{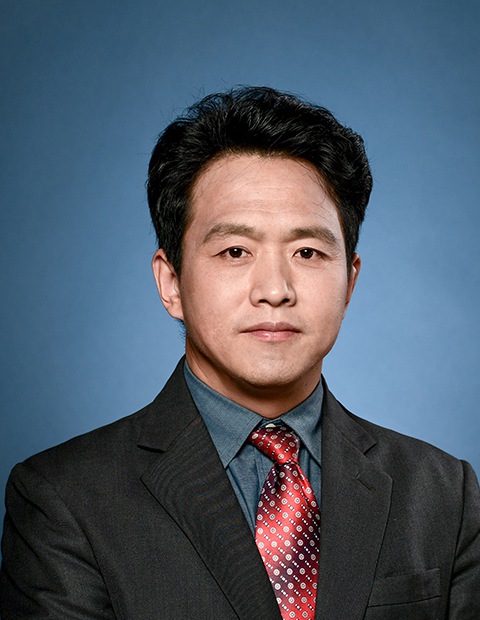}}]{Gongshen Liu}
received his Ph.D. degree in the Department of Computer Science from Shanghai Jiao Tong University in 2003. He is currently a professor with the School of Electronic Information and Electrical Engineering, Shanghai Jiao Tong University. His research interests cover natural language processing, machine learning, and artificial intelligence security.
\end{IEEEbiography}

\begin{IEEEbiography}[{\includegraphics[width=1in,height=1.25in,clip,keepaspectratio]{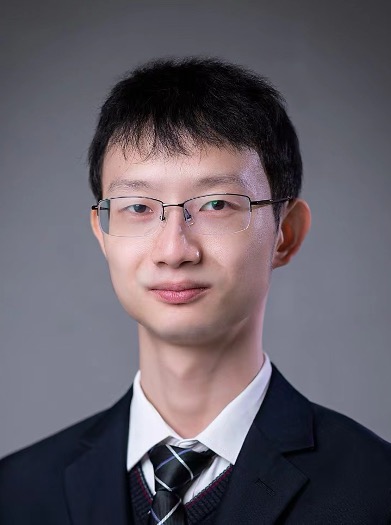}}]{Zhuosheng Zhang} received his Bachelor's degree from Wuhan University in 2016, and his M.S. and Ph.D. degrees from Shanghai Jiao Tong University in 2020 and 2023, respectively. He is currently a tenure-track assistant professor at Shanghai Jiao Tong University. He was a research intern at Amazon AWS, Microsoft Research, Langboat Technology, NICT (Japan), and IBM. His research interests include natural language processing, large language models, and language agents. 
He has published research papers in leading journals and conferences, such as TPAMI, TNNLS, TASLP, ICLR, ICML, ACL, AAAI, EMNLP, and COLING. 
He was the recipient of the WAIC 2024 Youth Outstanding Paper Award, WAIC 2024 YunFan Award, and the Global Top 100 Chinese Rising Stars in Artificial Intelligence. 
He serves as an action editor for ACL Rolling Review and standing reviewer for TACL. He served as a (senior) area chair for conferences such as NeurIPS, ACL, IJCAI, EMNLP, and COLING.
	\end{IEEEbiography}

\vfill

\end{document}